\documentclass[10pt,journal,compsoc]{IEEEtran}
\ifCLASSOPTIONcompsoc
  \usepackage[nocompress]{cite}
\else
  \usepackage{cite}
\fi

\ifCLASSINFOpdf

\else

\fi

\usepackage{graphicx}

\usepackage{comment}
\usepackage{amsmath,amssymb} 
\usepackage{color}
\usepackage{ulem}

\usepackage{bm}

\newcommand{\reversion}[1]{#1}

\usepackage{multirow}
\usepackage{mathrsfs}

\newcommand{\vect}[1]{{\bm{#1}}}

\usepackage{xspace}

\newcommand{\keep}[1]{}
\newcommand{\old}[1]{}

\DeclareMathOperator*{\softmax}{softmax}
\DeclareMathOperator*{\GCN}{GCN}
\DeclareMathOperator*{\CNN}{CNN}
\DeclareMathOperator*{\MLP}{MLP}

\DeclareMathOperator*{\FwdSkin}{SMPL}

\usepackage{times}
\usepackage{epsfig}
\usepackage{microtype}
\usepackage{tabu}                      
\usepackage{booktabs}                  
\usepackage{lipsum}                    
\usepackage{mwe}                       

\usepackage{mathptmx} 

\newcommand{\eqnref}[1]{{Equation~{#1}}}
\newcommand{\figref}[1]{{Figure~{#1}}}
\usepackage{hyperref}

\begin{document}

\title{Neural Novel Actor: Learning a Generalized Animatable Neural
Representation for \\Human Actors}

\author{
    Qingzhe Gao\textsuperscript{*},
    Yiming Wang\textsuperscript{*},
    Libin Liu\textsuperscript{†},
    Lingjie Liu\textsuperscript{†},
    Christian Theobalt,
    Baoquan Chen\textsuperscript{†}
    
\IEEEcompsocitemizethanks{
\IEEEcompsocthanksitem Qingzhe Gao is with the Department of Computer Science, Shandong University, China and Peking University, China. E-mail: gaoqingzhe97@gmail.com 
\IEEEcompsocthanksitem Yiming Wang, Libin Liu and Boquan Chen are with SIST \& KLMP (MOE), Peking University,China  E-mail: \{wym12416, libin.liu, baoquan\}@pku.edu.cn,
\IEEEcompsocthanksitem Lingjie Liu and is Christian Theobalt are with the Graphic, Vision \& Video group of Max Planck Institute for Informatics in Saarbrücken, Germany. E-mail:{\{lliu, theobalt\}@mpi-inf.mpg.de}
\IEEEcompsocthanksitem[*] The first two authors contributed equally to this work.
\IEEEcompsocthanksitem[†] These authors contributed senior supervision equally to this work.
}
}

\IEEEtitleabstractindextext{%
\begin{abstract}
We propose a new method for learning a generalized animatable neural human representation from a sparse set of multi-view imagery of multiple persons. The learned representation can be used to synthesize novel view images of an arbitrary person and further animate them with the user's pose control.
While most existing methods can either generalize to new persons or synthesize animations with user control, none of them can achieve both at the same time.
We attribute this accomplishment to the employment of a 3D proxy for a shared multi-person human model, and further the warping of the spaces of different poses to a shared canonical pose space, in which we learn a neural field and predict the person- and pose-dependent deformations, as well as appearance with the features extracted from input images.
To cope with the complexity of the large variations in body shapes, poses, and clothing deformations, we design our neural human model with disentangled geometry and appearance. Furthermore, we utilize the image features both at the spatial point and on the surface points of the 3D proxy for predicting person- and pose-dependent properties.
Experiments show that our method significantly outperforms the state-of-the-arts on both tasks.
\end{abstract}

}

\maketitle

\IEEEdisplaynontitleabstractindextext

\IEEEpeerreviewmaketitle

\IEEEraisesectionheading{
\section{Introduction}\label{sec:introduction}
}
\IEEEPARstart{S}{ynthesizing} high-quality free-viewpoint videos of an arbitrary human character using a sparse set of cameras is crucial for many computer graphics applications,  including VR/AR, film production, video games, and telepresence.
Many of these applications require  user control over human poses in the synthesis. 
Achieving these with traditional methods is difficult because 
it needs an expensive capturing setup \cite{debevec2000acquiring_array_c1,guo2019relightables_array_c2,carranza2003free_template1,de2008performance_template2,gall2009motion_template3,stoll2010video_template4}, the production-quality human geometry and appearance models, and manual invention and corrections \cite{collet2015high_dpeth1,dou2016fusion4d_depth2,su2020robustfusion_depth3}.

Recently, neural human representation and rendering algorithms based on the neural radiance fields (NeRF) \cite{mildenhall2020nerf_nerfbase} have demonstrated the ability to overcome the limitations of the traditional approaches. Some methods \cite{liu2021neuralactor,chen2021animatable,peng2021animatable,su2021nerf_A_nerf} can learn an animatable human representation from multi-view imagery in a person-specific setting, but they are not able to generalize to new persons. 
\reversion{
Other works \cite{kwon2021neural_gene_human1,zhao2021humannerf_gene_human2,kpnerf,chen2022gpNerf} proposed generalizable radiance fields for humans conditioned on input image features,} inspired by the generalized neural representation for static scenes \cite{yu2021pixelnerf_gene_nerf_0,wang2021ibrnet_gene_nerf_1,chen2021mvsnerf_gene_nerf3,chibane2021stereo_gene_nerf_4,liu2021neural_gene_nerf_5}. With the learned representations, they can generate novel views of an arbitrary person from sparse multi-view images without  training. However, their representations are not animatable and thus cannot generate images with user's pose control.

\begin{figure*}
  \centering
  \includegraphics[width=\linewidth]{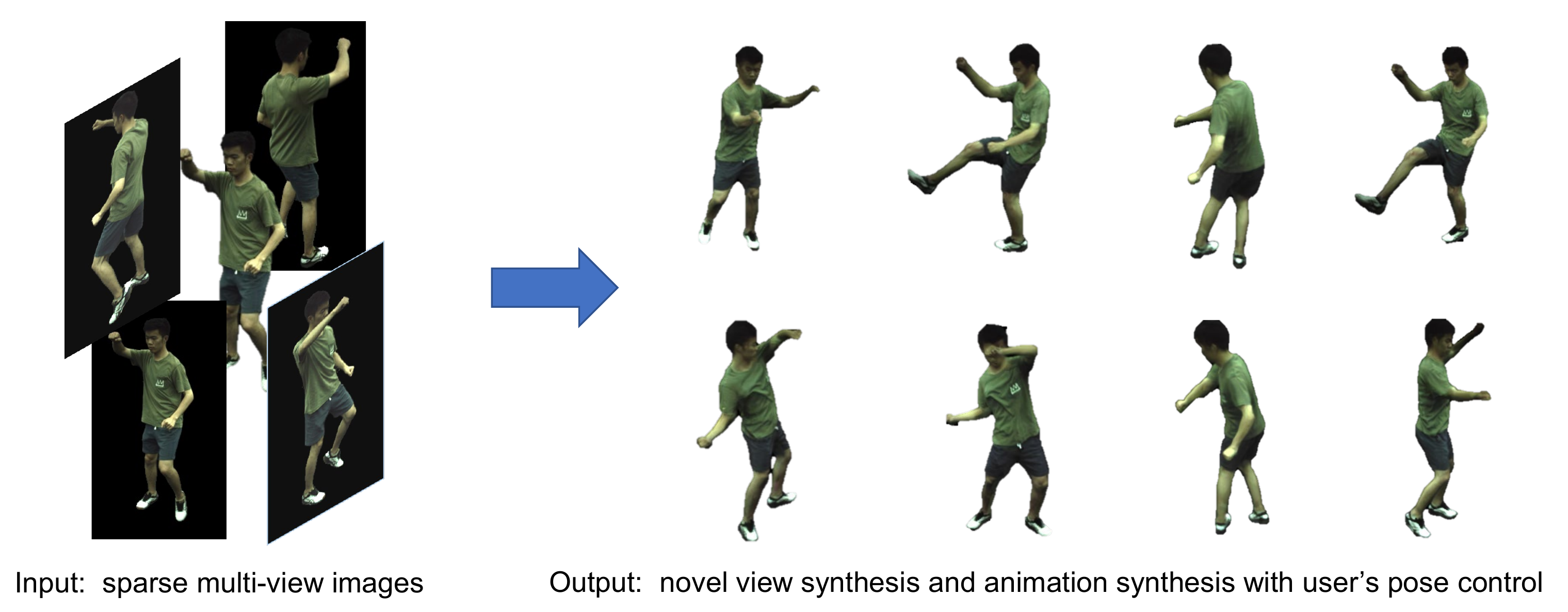}
  \caption{%
  Our model learns a generalized animatable neural human representation with multi-view RGB videos of several persons performing various motions.  At inference time, given sparse multi-view images, our model can directly get novel view synthesis and animation synthesis with user’s pose without further optimization.
  }
  \label{fig:teaser}
\end{figure*}

\reversion{In this paper, we address a challenging yet practical problem – training a model capable of rendering unseen individuals in a feed-forward manner using only still images captured from sparse viewpoints as input, as illustrated in Fig.~\ref{fig:teaser}.} To tackle this task, 
we introduce a new approach to learning a generalized animatable neural human representation from sparse multi-view input imagery of multiple persons. This representation allows us to generalize to new persons without training and further animate the representation with pose control. 

Specifically, our method uses a Skinned  Multi-Person Linear (SMPL) model as a 3D proxy and transforms each pose space to a shared canonical pose space. Then a neural radiance field in the canonical space is learned and we estimate person- and pose-dependent geometry and appearance with the features extracted from the input images. To efficiently learn this representation for multiple persons, we disentangle the geometry and appearance in our human model by extracting separate features for geometry and appearance properties to condition the prediction of these properties. Furthermore, we extract the image features at both the spatial points and the surface points of SMPL to better infer the person- and pose-dependent properties. 

We evaluate our method on the ZJU-MoCap \cite{peng2021neural_human_nerf_2}, DeepCap \cite{habermann2020deepcap} and DynaCap \cite{habermann2021real_DynaCap} datasets. 
\reversion{Our method significantly outperforms the state-of-the-art, MPS-NeRF\cite{gao2022mps}, in this challenging task. To demonstrate the effectiveness of our representation, we also separately evaluate its performance in two key aspects: generalization of novel view synthesis and animation with user-defined pose control. The experiments show that our method outperforms the state-of-the-art on both these tasks. These evaluations provide compelling evidence that our method has the potential to serve as a robust baseline model for future research endeavors in the domain of generalizable 3D human rendering.}

In summary, our technical contributions are:
\begin{itemize}
    \item We present a new method for achieving both the novel view synthesis of arbitrary persons and the animation synthesis with pose control at the same time. 
    \item We design a new generalized animatable neural human representation with disentangled geometry and appearance, which can be learned efficiently from sparse multi-view imagery of multiple persons. 
    \item We present a new way to predict the person- and pose-dependent properties by taking the features at both the spatial points and the surface points of SMPL into account. 
\end{itemize}

\section{Related Work}
\subsection{Human Performance Capture}
There have been multiple studies addressing novel view synthesis of human performance. While many methods based on pre-scanned human models \cite{carranza2003free_template1,de2008performance_template2,gall2009motion_template3,stoll2010video_template4} can capture humans in a sparse multi-view setting, pre-scanned human models are often unavailable in most cases.
Recent works rely on depth sensors \cite{collet2015high_dpeth1,dou2016fusion4d_depth2,su2020robustfusion_depth3} or dense arrays of cameras \cite{debevec2000acquiring_array_c1,guo2019relightables_array_c2} to achieve high-fidelity reconstruction, but these settings are not easily accessible. By employing neural networks, some methods \cite{martin2018lookingood_net_c1,wu2020multi_c2,meshry2019neural_c3} can compensate for geometric artifacts through the modification of the rendering pipeline.
More recently, several works \cite{saito2019pifu_single2,natsume2019siclope_single1,rahaman2019spectral_single3,zheng2019deephuman_single4,saito2020pifuhd_single5,huang2020arch_single6,he2021arch++_single7} have been able to reconstruct 3D humans from a single image using 3D human geometry priors. However, these methods rely on 3D geometry data and cannot generalize to complex poses that are not present in the training data. In contrast, our method is capable of generalizing to new persons using only sparse multi-view image supervision.

\subsection{Neural Representations for human}
Neural rendering techniques \cite{shysheya2019textured_nppl_0,thies2019deferred_nppl_1,liu2020neural_nppl_2,wu2020multi_c2,kwon2020rotationally_nppl_4} have enabled neural networks to learn 3D object reconstruction from 2D images. Various 3D representations, such as 3D voxel-grid \cite{lombardi2019neural_voxel_1,sitzmann2019deepvoxels_voxel_2,kwon2020rotationally_voxel_5,yan2016perspective_voxel_6}, point clouds \cite{aliev2020neural_point_1,wu2020multi_c2}, textured mesh \cite{liu2019neural_texture_1,liao2020towards_texture_3,xiang2021neutex_texture_4,thies2019deferred_nppl_1}, and multi-plane images \cite{flynn2019deepview_mpl_1,zhou2018stereo_mpl_2,tucker2020single_mpl_3}, have been learned from 2D images through differentiable rendering to enhance novel view synthesis performance. However, achieving higher resolution remains challenging due to memory constraints.

NeRF \cite{mildenhall2020nerf_nerfbase} represents scenes with implicit fields of density and color. To extract more accurate surfaces, some works \cite{yariv2020IDR,DVR2020,yariv2021volume_volumesdf,wang2021_neus} employ the signed distance function (SDF) to represent geometry in a scene. Building on these representations, numerous studies \cite{gao2020portrait_human_nerf_1,peng2021neural_human_nerf_2,pumarola2021d_human_nerf_3,chen2021moco_human_nerf_5,xian2021space_human_nerf_4,noguchi2021neural_human_nerf_6,su2021nerf_A_nerf,singHumanNeRF,xu2022surface} use neural representation to capture humans. However, optimizing for each novel video is time-consuming.

\reversion{Generalizable neural representation methods \cite{yu2021pixelnerf_gene_nerf_0,wang2021ibrnet_gene_nerf_1,chen2021mvsnerf_gene_nerf3,chibane2021stereo_gene_nerf_4,liu2021neural_gene_nerf_5, yao2021ddnerf} address this issue by employing implicit fields conditioned on image features. Inspired by these works, some studies \cite{kwon2021neural_gene_human1,zhao2021humannerf_gene_human2,kpnerf,chen2022gpNerf} propose generalizable radiance fields for humans, but they fail to achieve an animatable human model.}

Leveraging the Skinned Multi-Person Linear (SMPL) model \cite{loper2015smpl_SMPL}, several works \cite{liu2021neuralactor,chen2021animatable,peng2021animatable,su2021nerf_A_nerf, hdhumans} manage to obtain an animatable neural representation for humans. However, they still require subject-specific training. In contrast, our proposed novel deformable neural human representation enables us to acquire an animatable neural human representation from a single multi-view image of a new person without the need for additional training.

MPS-NeRF \cite{gao2022mps} can also learn an animatable neural human representation from multi-view images of a single frame of the target person. 
It mainly relies on inverse skinning weights of SMPL\cite{loper2015smpl_SMPL} to animate the representation, while our method  learns an additional
residual deformation mapping to compensate
for the deformation that cannot be modeled by inverse kinemetric transformation. Furthermore, the representation of MPS-NeRF is based on NeRF \cite{mildenhall2020nerf_nerfbase} , while  our representation additionally disentangles geometry and appearance by formulating them as two separate template implicit functions based on NeuS \cite{wang2021_neus}.

\begin{figure*}
    \centering
    \includegraphics[width=0.95\textwidth]{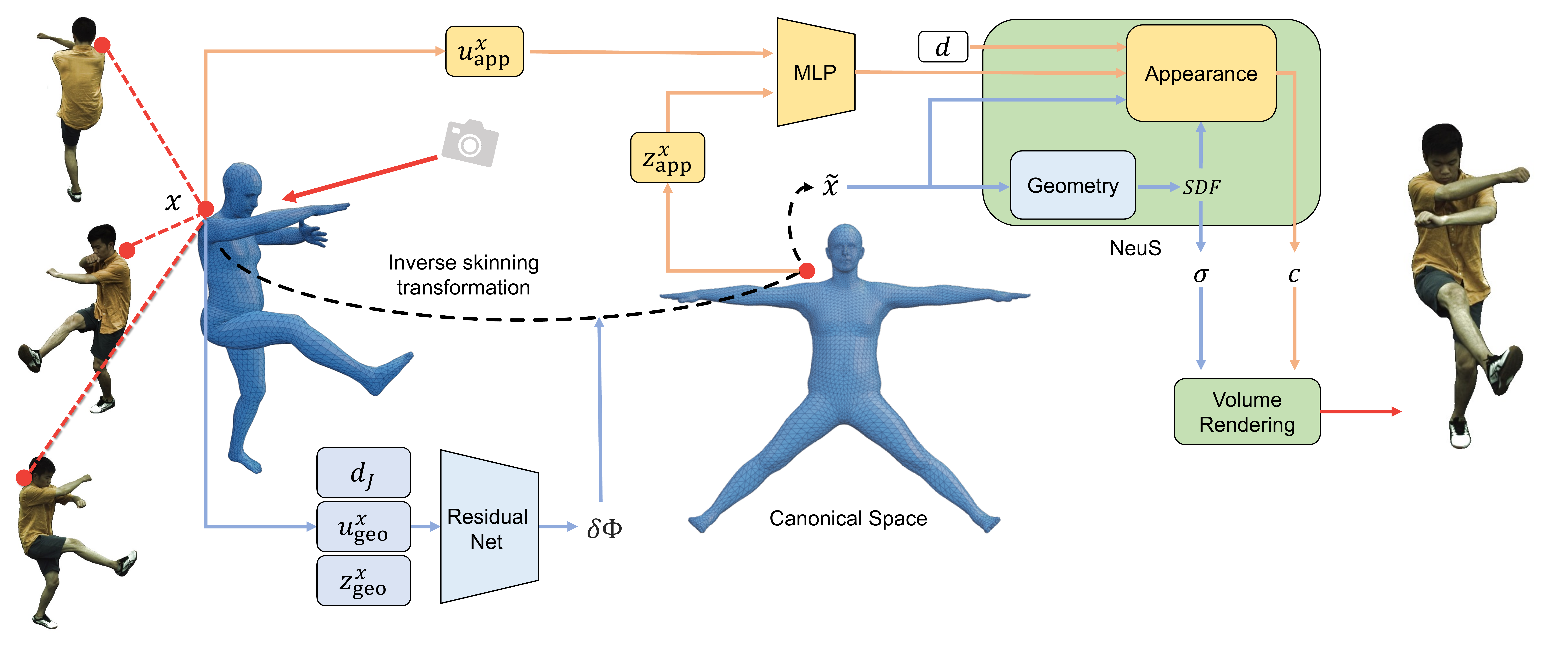}
    \caption{Overview of our framework. Given a query point $\vect{x}$ in the posed space, we use inverse skinning transformation of its nearest surface point and a predicted residual deformation $\delta\Phi$ to transform $\vect{x}$ to the canonical space. The deformed point $\tilde{\vect{x}}$ is used as input to our geometry network and appearance network. The pose-dependent residual deformation is predicted using geometry features $\vect{u}^{{x}}_{\text{geo}},\vect{z}^{{x}}_{\text{geo}}$ and the relative displacement $\vect{d}_J$ between the query point $\vect{x}$ and every joint. The appearance features $\vect{u}^{{x}}_{\text{app}},\vect{z}^{{x}}_{\text{app}}$ are used as input to the appearance network. The appearance network also takes the view direction $\vect{d}$ as input.}
    \label{fig:system_overview}
\end{figure*}

\section{Method}
Given a set of multi-view RGB videos of several persons performing various motions, our goal is to learn a generalized animatable neural human representation (Fig. {\ref{fig:system_overview}}) in the training.
At inference time, our model enables two tasks:
(1) Generalization: given a sparse multi-view (e.g., 3 or 4 views) videos of an unseen person performing arbitrary motions, we can synthesize novel views of the person performing these motions without training.
(2) Animation: given a sparse multi-view images of an unseen person in a static pose, we can animate the neural representation of the person to generate novel pose images according to the user's pose control.

Our method uses the Skinned Multi-Person Linear (SMPL) model \cite{loper2015smpl_SMPL} as a 3D proxy and learns a canonical pose space shared by all the persons and poses. 
For each 3D point in a posed space, we convert it into this canonical space using the inverse skinning transformations \cite{huang2020arch_single6} and the non-rigid deformations predicted by a neural network. Then, we learn a neural field \cite{wang2021_neus} in the canonical space to infer Signed Distance Fields (SDF) and color for generating the final images.
Our key idea is to extract geometry and appearance features for each 3D point from the sparse input images and use these features to infer the non-rigid residual deformations, SDF and color at the 3D point. 
To better train our model on multiple persons, we propose two new designs: 
(1) We disentangle geometry and appearance by formulating them as two separate template implicit functions in the canonical space.
(2) We extract the geometry and appearance features at both the 3D point and its nearest SMPL vertices; the rationale is that our model is a person-agnostic model and the 3D proxy (i.e., SMPL model) for different persons is shared. Therefore, taking the  properties at both the 3D point and its SMPL surface points into consideration would better infer the distance of the 3D point to the SMPL surface for different persons.

In the following, we first introduce our deformable neural human representation 
(Sec. \ref{deformable_representation}) 
and then explain how we construct the geometry and appearance features used in such a representation (Sec. \ref{construction_features}). Based on our neural representation, we can synthesize novel view images for arbitrary human poses 
(Sec. \ref{pose_driven_rendering}).

\subsection{Deformable Neural Human Representation} \label{deformable_representation} 
In order to represent different human identities and poses, we adopt SMPL ~\cite{loper2015smpl_SMPL} as the base model in our framework. 
SMPL is a mesh-based human model consisting of a template mesh with $N_v=6890$ vertices $V\in\mathbb{R}^{N_v\times{}3}$ and driven by $N_J$ joints.
It deforms the template mesh according to a set of parameters, $\vect{\rho}$, representing the body shape and pose of a person. This process can be written as
\begin{align}
    \vect{v}_{\bm{\rho}}=\FwdSkin\big(\vect{v},\bm{\rho},\vect{w}_v\big),
    \label{eqn:fwd_skin}
\end{align}
where $\vect{v}\in{}V$ represents a surface vertex and $\vect{w}_v$ is the skinning weight of $\vect{v}$. 
We consider a predefined pose $\vect{\rho}_0$ as the {canonical model} of our framework. The {canonical space} is then defined as the corresponding 3D space containing this canonical model.
For an arbitrary pose $ \vect{\rho}$, we can transform the spacial points in the corresponding {posed} space into this canonical space using the SMPL surface as guidance. 
Such transformation links the canonical space to different posed spaces and thus allow those spaces to share the common features defined in the canonical space.



\subsection{Geometry and appearance features}
\label{construction_features}

Our framework utilizes two latent variables, $\vect{F}_{\text{geo}}$ and $\vect{F}_{\text{app}}$, to represent the geometry and appearance features of a person, respectively. 
These features are extracted from the input images and then used to render new images of the person performing novel poses from novel views.
Our framework obtains the geometry and appearance features in a similar manner, so we omit the subscripts in this section for simplicity when it is not confusing.

Both the geometry and appearance features are defined at every spacial point around the human model. The latent variable $\vect{F}$ defined at a spacial point $\vect{x}$ consists of two components $\vect{F}=(\vect{u}^{{x}},\vect{z}^{{x}})$, where $\vect{u}^{{x}}$ represents the image features based on pixel alignment, and $\vect{z}^{{x}}$ are surface features computed based on the connectivity of the SMPL mesh.

\subsubsection{Occlusion-aware image features}
Formally, our framework takes a sparse set of multi-view images, $\{I^c\}, c=1,\dots,N_C$, captured by $N_C$ calibrated cameras as input, where each image $I^c\in\mathbb{R}^{H\times{}W\times{}(3+1)}$ contains the RGB color of every pixel and a foreground mask indicating the pixels belong to the person.
We apply CNN~\cite{krizhevsky2012imagenet_CNN} to these images and extract a set of feature maps $\bm{U}^c$ at multiple resolutions as
\begin{align}
    \bm{U}^{c} = \CNN([I^c]).
\end{align}
Then, for every spacial point $\vect{x}$, we obtain a set of image features $\{\vect{u}^c\}$ by projecting $\vect{x}$ onto each feature map $\bm{U}^{c}$.
Inspired by \cite{kwon2021neural_gene_human1}, we employ a self-attention mechanism to aggregate these image features from different views to compute the image features $\vect{u}^x$ at $\vect{x}$ as
\begin{align}
    \label{eqn:occ_bias_att}
    \bm{u}^x= \softmax_{c} \left(\frac{Q(\bm{u}^c) \cdot K(\bm{u}^c)}{\sqrt{d}} + B^c \right) \cdot V(\bm{u}^c),
\end{align}
where $Q(\cdot)$, $K(\cdot)$, and $V(\cdot)$ are the learnable query, key, and value embedding functions proposed in the original self-attention mechanism~\cite{NIPS2017_attention}, and $d$ is the dimension of the embedding space. 

When a spacial point is occluded in an input image $I^c$, its corresponding image features $\vect{u}^c$ is not reliable and should not be weighted the same as the features from the other images. We thus employ an occlusion-aware mechanism to ensure this principle. Specifically, when a 3D query point $\vect{x}$ is occluded by the posed SMPL model in an input image $I^c$, we subtract a fixed bias $B^c$ from the corresponding attention weights in \eqnref{\eqref{eqn:occ_bias_att}} to explicitly inform the self-attention mechanism of this occlusion. This mechanism is partially inspired by Attention with Linear Biases (ALiBi) \cite{Press2021TrainST}. We find that it effectively mitigates the artifacts caused by occlusion in our experiments. 



\subsubsection{Pose-aware surface features} \label{skel_feature}
The mesh structure of the SMPL model provides a strong prior for determining the shape and appearance of human body. To fully utilize such structural cues, we associate surface features to the mesh and diffuse them to the spacial points in the surrounding space.

To build these surface features, we extract the occlusion-aware image features of every vertex of the mesh from the image and employ the Graph Convolution Networks (GCN) to fuse them as suggested by \cite{GCN_original,pfaff2020mesh_gcn,sanchez2020gcn_simulate}.
GCN is a special convolutional network structure that aggregates the information on each individual vertex based on the connectivity of the mesh. We additionally include the displacement between each pair of connected vertices in this operation. Considering that the SMPL mesh is deformed based on pose parameters, this augmentation effectively allows the GCN to encode an implicit pose description into the computation, thus resulting in a set of pose-aware features. 

\reversion{To enhance the generalization ability of GCN, we use a novel local representation instead of global pose parameters (e.g. SMPL's 72 dimension of pose vector) as the input pose information of GCN. Specifically, the pose information is included in the graph edges, by using the direction and length of the posed edges as their features. The edge features are scattered to the node features to conduct message passing for the GCN. This localized pose representation helps the GCN to generate pose-dependent appearance and geometry to reduce reconstruction loss during training and generalize to new poses after training on a dataset with a large variety of poses.}


Formally, assuming the input images correspond to pose parameters ${\vect{\rho}_I}$, we compute a deformed mesh $V_{{\rho}_I}$ using SMPL and project each vertex $\vect{v}\in{}V_{{\rho}_I}$ onto the input images, obtaining a set of occlusion-aware image features $\{\vect{u}^v\}$. Then, we convert these image features using the pose-aware GCN described above and compute surface features
\begin{align}
    \{{\vect{z}}^v_I\} = \GCN(\{\vect{u}^v\},V_{{\rho}_I}).
    \label{eqn:de_pose_gcn}
\end{align}
When rendering a new pose $\vect{\rho}$, we transform these input surface features $\{{\vect{z}}^v_I\}$ onto the corresponding new SMPL mesh $V_{\vect{\rho}}$ via another GCN procedure
\begin{align}
    \{{\vect{z}}^v\} = \GCN(\{\vect{z}^v_I\},V_{{\rho}}).
    \label{eqn:pose_gcn}
\end{align}
\reversion{The GCNs of \eqnref{\eqref{eqn:de_pose_gcn}} and \eqnref{\eqref{eqn:pose_gcn}} do not share weights.}
Note that these surface features, $\{{\vect{z}}^v\}$, are defined only on the surface vertices $\vect{v}\in{}V_{\rho}$. We then extend them to the surrounding space through a diffusion process. Specifically, for a spatial point $\vect{x}$ in the vicinity of the deformed mesh $V_{\rho}$, we find $K$ nearest vertices $\{\vect{v}^k\}\subset{}V_{\rho}$ on the mesh and take their features $\{\vect{z}^{k}\}$. The surface feature $\vect{z}^x$ of the query point $\vect{x}$ is then computed as 
\begin{equation}
    \vect{z}^x=\sum_k w_k \MLP(\vect{z}^{k}, \vect{x} - \vect{v}^k),
\end{equation}
where $w_k=({\lVert \vect{x} - \vect{v}^k  \rVert+\epsilon})^{-1}/\sum_k({\lVert \vect{x} - \vect{v}^k  \rVert+\epsilon})^{-1}$, and $\epsilon$ is a small scalar used to prevent dividing by zero. 

\subsubsection{Implementation} 
For a spacial point $\vect{x}$, the latent variable $\vect{F}_{\text{geo}}=(\vect{u}^{{x}}_{\text{geo}},\vect{z}^{{x}}_{\text{geo}})$ and $\vect{F}_{\text{app}}=(\vect{u}^{{x}}_{\text{app}},\vect{z}^{{x}}_{\text{app}})$ are computed using the images features and the surface features defined above. To further enforce disentanglement of the features, we let the appearance features $\vect{F}_{\text{app}}$ be independent of the driving pose $\vect{\rho}$. This is achieved by computing $\vect{z}^{{x}}_{\text{app}}$ using the canonical pose $\vect{\rho}_0$ in \eqnref{\eqref{eqn:pose_gcn}}, as depicts in \figref{\ref{fig:system_overview}}.

\subsection{Pose-driven Volume Rendering} \label{pose_driven_rendering}
Using the geometry and appearance features $\vect{F}_{\text{geo}}$ and $\vect{F}_{\text{app}}$ extracted from the input images, our framework can render new images from a novel viewpoint given an arbitrary driving pose $\vect{\rho}$. We employ NeuS \cite{wang2021_neus}, an SDF-based differential renderer, to synthesize those images.
NeuS predicts the color of each pixel by accumulating the radiance along the camera ray $\vect{r}$ passing through the pixel. This computation can be discretized using a series of spacial points $\{\vect{x}_i\}$ sampled along $\vect{r}$. Specifically, NeuS computes
\begin{align}
    \tilde{C}(\vect{r}) = \sum_{i=1}^{n}T_i\alpha_i{c}_i ,
\end{align}
where $\tilde{C}(\vect{r})$ is the predicted color, $T_i=\prod_{j=1}^{i-1}(1-\alpha_i)$ is the discrete accumulated transmittance, $\alpha_i$ represents the opacity values defined as
\begin{align}
    \alpha_i = \max\left(\frac{\phi(s_i)-\phi({s}_{i+1})}{\phi({s}_i)},0\right),
\end{align}
and $\phi(x)=(1+e^{-kx})^{-1}$ with a learnable scalar $k$. The color values $c_i$ and the SDF values $s_i$ in the above equations are evaluated at every sample point $\vect{x}_i$. Our framework handles every sample point in the same way, so we omit the subscript $i$ in the rest of this section for simplicity.

\subsubsection{Pose-driven deformation field}
Given a driving pose $\vect{\rho}$, we transform each sample point $\vect{x}$ into the canonical space and evaluate $c$ and $s$ based on the canonical position $\tilde{\vect{x}}$. This mechanism is inspired by recent studies \cite{Tretschk2021_NonRigid,liu2021neuralactor}, which have shown its efficiency in modeling dynamic scenes and human poses. We define the deformation mapping $\Phi$ using the inverse skinning transformation \cite{huang2020arch_single6}. As suggested by \cite{liu2021neuralactor}, an additional residual deformation mapping $\delta\Phi$ is employed to compensate for the deformation that cannot be captured by the inverse skinning, such as the deformation of cloth. The canonical position $\tilde{\vect{x}}$ of a sample point $\vect{x}$ is thus computed as 
\begin{equation}
    \tilde{\vect{x}} = \Phi(\vect{x},\vect{\rho}) + \delta\Phi(\vect{x},\vect{\rho}).
    \label{eqn:deform_field}
\end{equation}

$\Phi(\vect{x},\vect{\rho})$ is the inverse skinning mapping
\begin{equation}
    \Phi(\vect{x},\vect{\rho}) = \sum_{j=1}^{N_J} w_j \cdot
    \left(R_j(\vect{x}-\delta{}\vect{v}) + \vect{t}_j\right),
    \label{eqn:bwd_skin}
\end{equation}
where $R_j$ and $\vect{t}_j$ represent the rotation and translation that transform joint $j$ from pose $\vect{\rho}$ back to the canonical pose $\vect{\rho}_0$, $w_j\in\vect{w}_v$ is the corresponding skinning weight of the surface point $\vect{v}$ that is the nearest to $\vect{x}$, and $N_J$ is the number of joints.
Note that we allow the pose parameters to also define the body shape of the target person. A displacement $\delta\vect{v}$ is leveraged to compensate for the deformation caused by the change of body shape. Specifically,
\begin{equation}
    \delta{}\vect{v} = \FwdSkin\big(\vect{v},\beta(\vect{\rho}),\vect{w}_v\big)-\FwdSkin\big(\vect{v},\beta(\vect{\rho}_0),\vect{w}_v\big),
\end{equation}
where $\beta(\cdot)$ extract the body shape parameters from $\vect{\rho}$.

The residual deformation $\delta\Phi$ is computed using the geometry features $\bm{F}_{\text{geo}}^x$ extracted from the input images. 
We further consider the relative displacement between the query point $\vect{x}$ and every joint, collectively represented by $\vect{d}_J$, as an extra cue. The residual displacement is thus computed as 
\begin{equation}
    \delta\Phi(\vect{x},\vect{\rho}) = \MLP(\vect{F}_{\text{geo}}^x,\vect{d}_J).
\end{equation}

\subsubsection{SDF-based volume rendering} 
After transforming the sample points $\vect{x}$ into the canonical space $\tilde{\vect{x}}$ using the deformation field, we compute its SDF value $s$ and color $c$ as
\begin{align}
    s& = {\mathcal{S}}(\tilde{\vect{x}}) \\
    c& = {\mathcal{C}}(\tilde{\vect{x}}, \vect{F}_{\text{app}},\vect{d},s,\vect{n}_x).
    \label{eqn:color_func}
\end{align}
Both ${\mathcal{S}}$ and ${\mathcal{C}}$ are implemented as MLPs. 
The SDF function ${\mathcal{S}}$ only takes the canonical position of $\vect{x}$ as input.
%
The color function $\mathcal{C}$ considers the appearance feature $\bm{F}_{\text{app}}$, the view direction $\vect{d}$, the SDF value $s$ of the sample point, as well as 
the normal vector $\vect{n}_{{x}}$ of the implicit surface at the sample point. $\vect{n}_{{x}}$ can be computed as the gradient of the SDF function $\vect{n}_{{x}}=\nabla_x{\mathcal{S}}(\vect{x})$.
The results of these functions are then used by NeuS to predicts the color of the pixel as described above.




\begin{figure*}[t]
    \centering
    \includegraphics[width=0.95\linewidth]{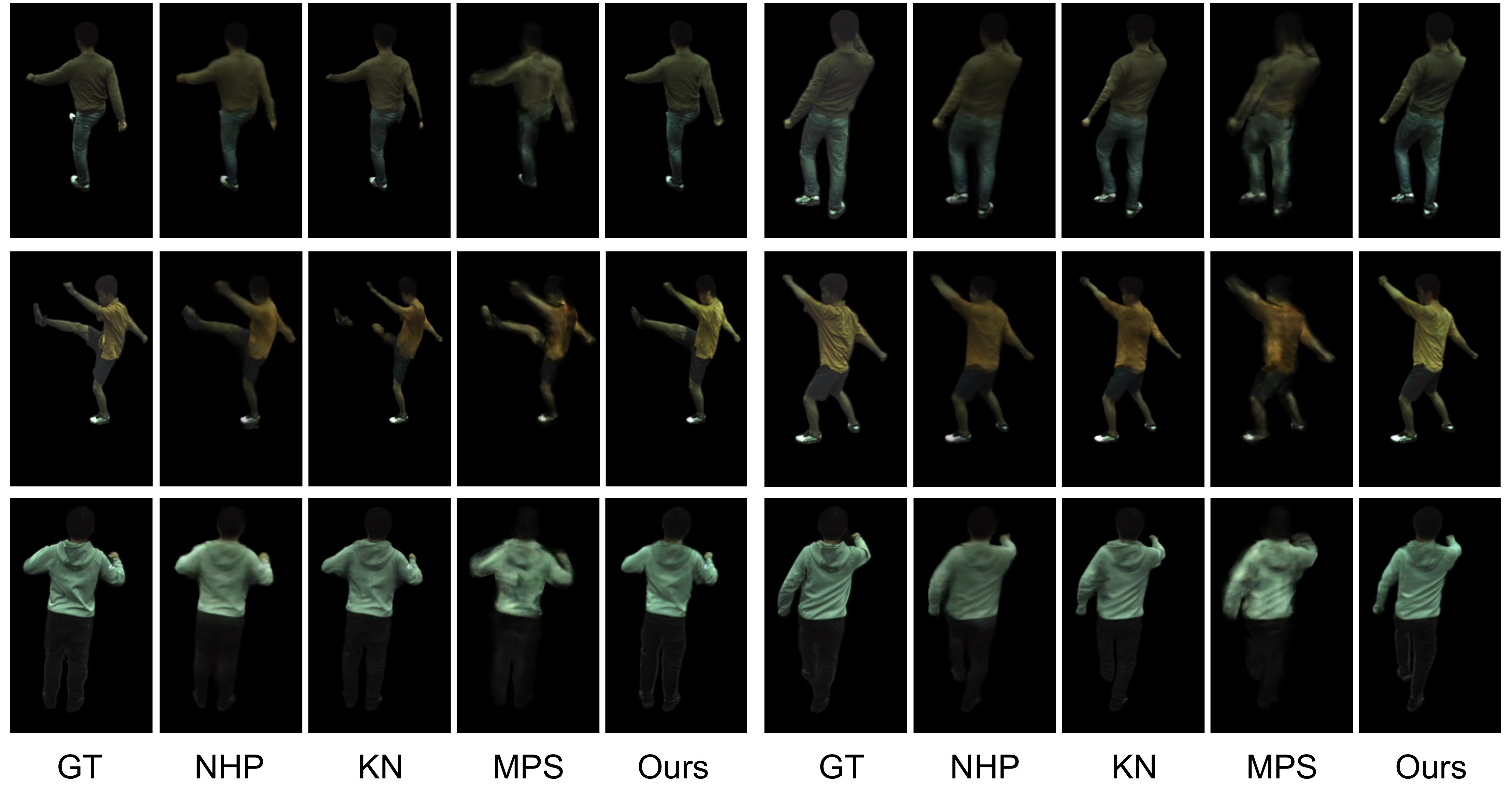}
    \caption{
    \reversion{Qualitative comparison of identity generalization on the ZJU-MoCap~\cite{peng2021neural_human_nerf_2} dataset. Our method outperforms three baselines, Neural Human Performer (NHP)~\cite{kwon2021neural_gene_human1}, Keypoint NeRF (KN)~\cite{kpnerf} and MPS-NeRF (MPS)~\cite{gao2022mps} in terms of synthesized wrinkles and appearance details. All methods are \textbf{trained on all the source subjects} and directly \textbf{tested on the target subjects} without training.
    } 
    } \label{fig:unseen_people}
\end{figure*}

\begin{figure*}[t]
    \centering
    \includegraphics[width=0.95\linewidth ]{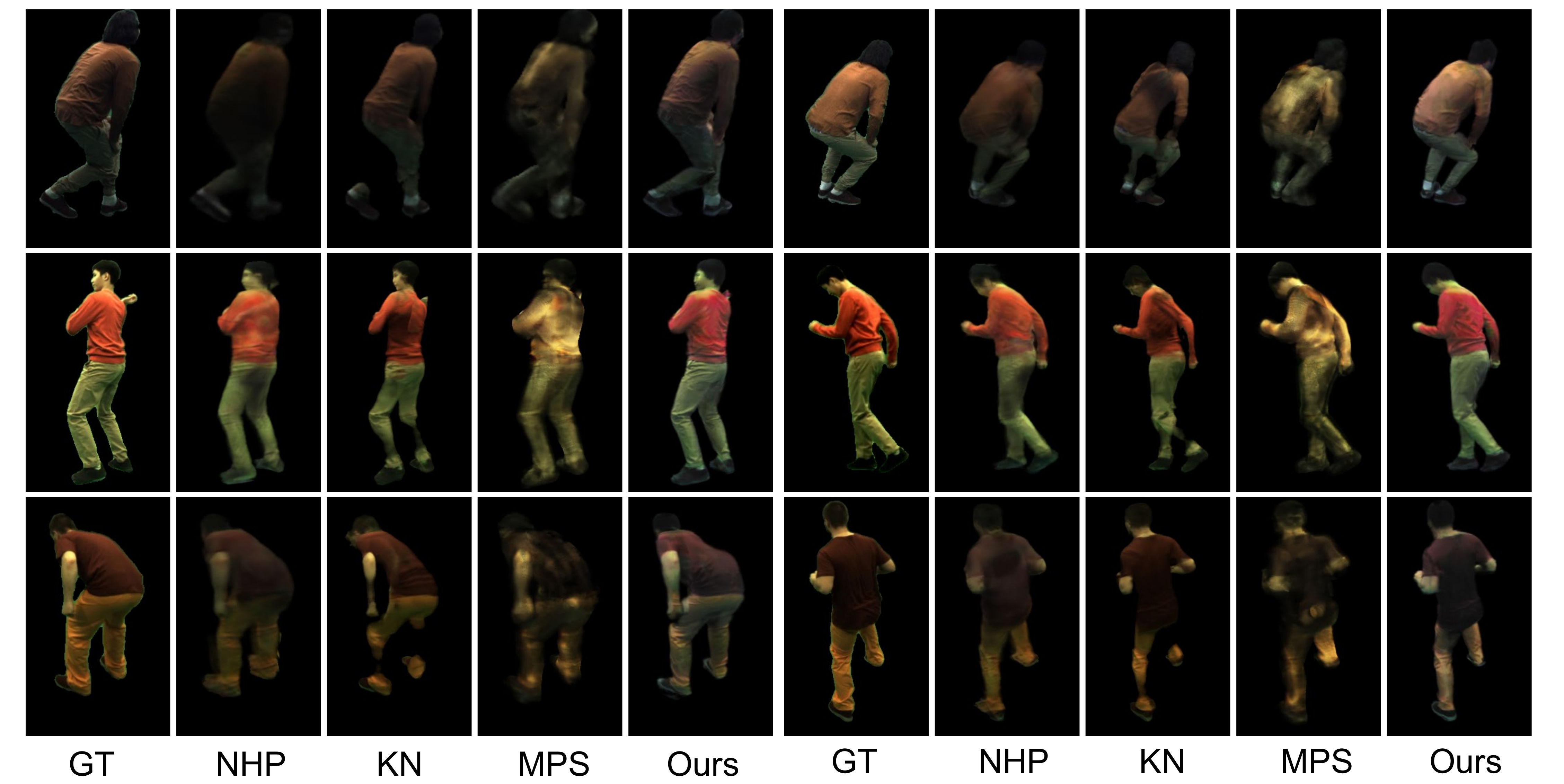}
    \caption{Qualitative comparison of cross-dataset generalization on novel view synthesis. All methods are trained on the ZJU-MoCap~\cite{peng2021neural_human_nerf_2}  and directly tested on the DeepCap~\cite{habermann2020deepcap} and DynaCap~\cite{habermann2021real_DynaCap}. Our method significantly outperforms other baselines. }\label{fig:cross_dataset}
\end{figure*}

\subsection{Training}

For every training image, we render $m$ random pixels and sample $n$ spacial points on each generated camera ray. The loss function is then defined as
%
\begin{equation} \label{eqn:loss}
\begin{split}
       \mathcal{L} = \frac{1}{m}\sum_{r} \underbrace{\Vert{\tilde C(r)-C(r)}\Vert_1}_{\text{color \ loss}}
       +\lambda_1 \sum_{r}\underbrace{\text{BCE}(\tilde M_r,M_r)}_{\text{mask loss}}  \\
       + \lambda_2 \frac{1}{nm}\sum_{\bm{x}} 
       \underbrace{(\Vert\bm{n}_{\bm{x}}\Vert_2-1)^2}_{\text{eikonal term}}
       +\lambda_3 \underbrace{\text{LPIPS}(\tilde C(P), C(P))}_{\text{LPIPS loss}},
\end{split}
\end{equation}
where the color loss measures the difference between the predicted color $\tilde C(r)$ and the ground truth $C(r)$, the mask loss matches the predicted mask $\tilde M_r = \sum_{i=1}^{n}T_i\alpha_i$ with the foreground mask $M_r$ by computing the binary cross entropy (BCE) between them, and an Eikonal term~\cite{gropp2020igr} is adopted to regularize the SDF function. 
We further employ a perceptual loss, LPIPS \cite{zhang2018perceptual}, to ensure the quality of the synthesized image. This LPIPS loss is computed by rendering random patches $P$ sampled on the target image. 
\begin{figure*}[t]
    \centering
    \includegraphics[width=0.95\linewidth ]{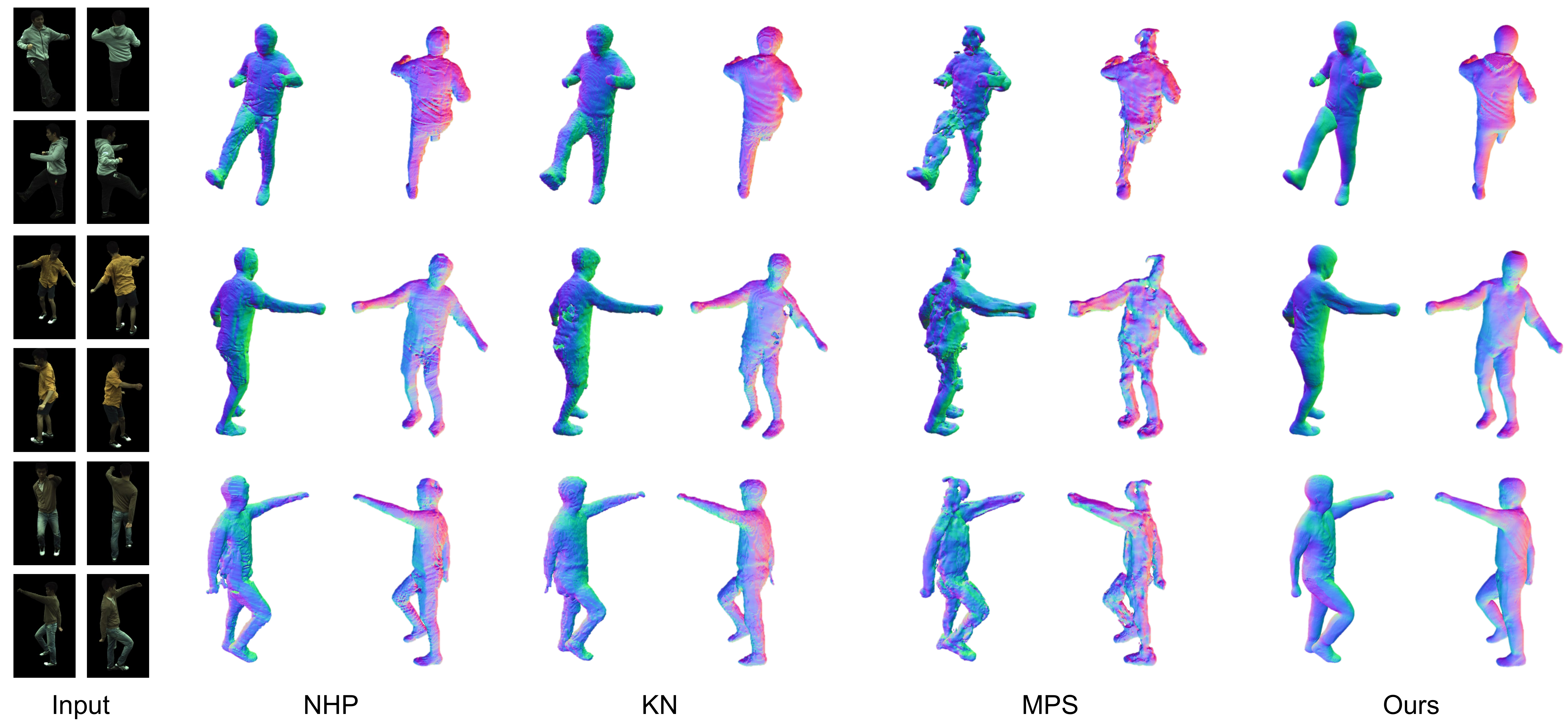}
    \caption{Visualization of 3D reconstruction results. The meshes are extracted by running the Marching Cubes algorithm on the predicted volume density.
  Other methods contain more unwanted artifacts compared to our method.
    }\label{fig:mesh_rusults}
\end{figure*}

\reversion{During training, we utilized multi-view inputs with multiple frames to introduce pose variation. In detail, we randomly selected a frame of a character and used its multi-view inputs to train the model. This involved reconstructing an image of another view using any three available views and calculating the loss. During testing, we only use a single frame of multiview image to generate an animatable avatar, which is consistent with our training setting.
The process of obtaining image features to render a new pose involves transforming a 3D point from the new pose to the input pose to acquire image features, using the canonical space as a bridge.
Besides image features, we also get pose-aware surface features for the new pose using the GCN as described.}

\begin{figure*}[t]
    \includegraphics[width=0.95\linewidth]{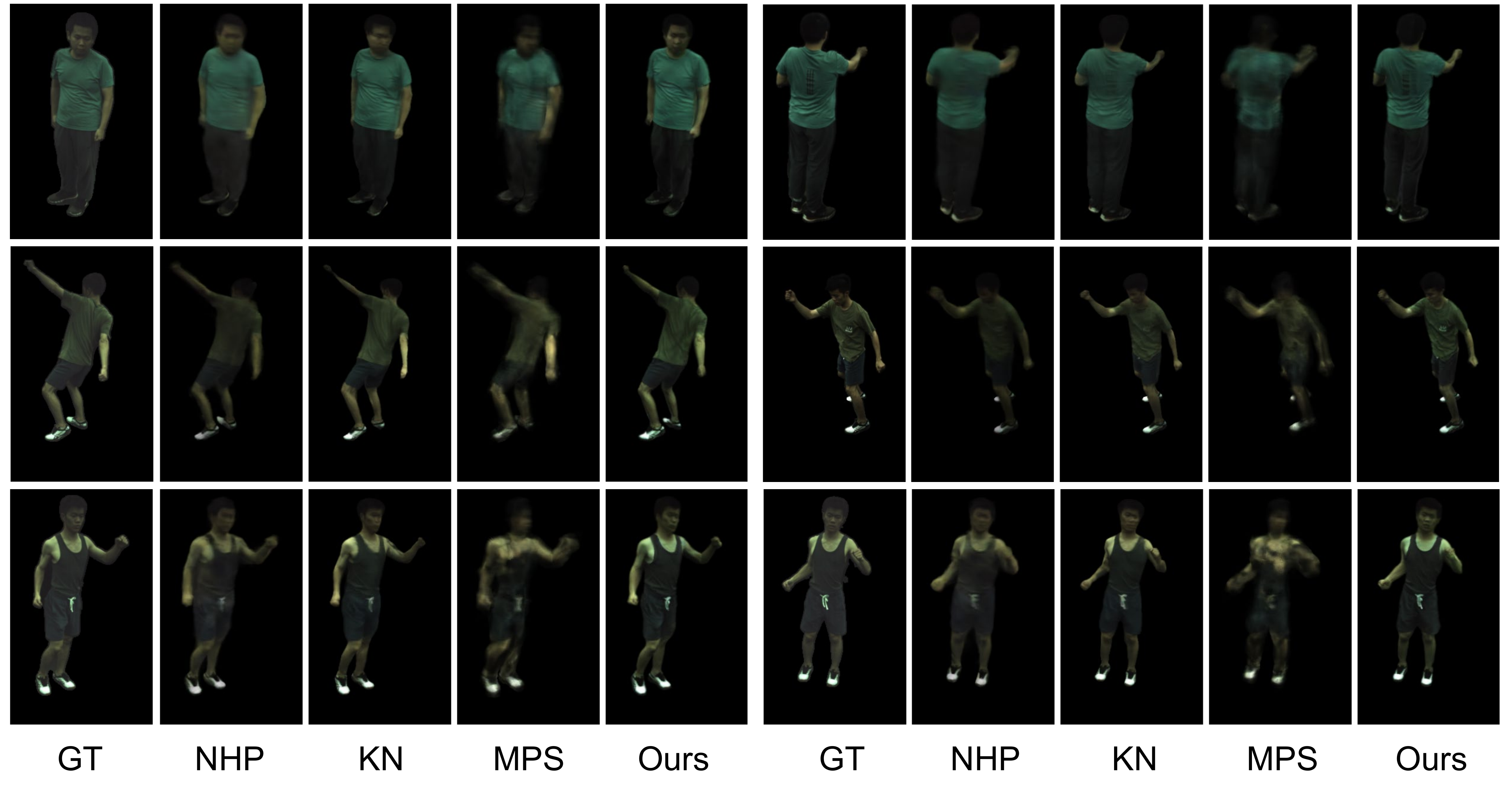}
    \caption{
    \reversion{
    Qualitative comparison of Pose generalization on the ZJU-MoCap dataset. Our method significantly outperforms three baselines,  Neural Human Performer (NHP)~\cite{kwon2021neural_gene_human1}, Keypoint NeRF (KN)~\cite{kpnerf} and MPS-NeRF (MPS)~\cite{gao2022mps}, in terms of synthesized wrinkles and appearance details.
    }
    }\label{fig:unseen_pose}
\end{figure*}


\section{Experiment}
\textbf{Implementation details.} We train our models using the Adam~\cite{Adam} optimizer. We follow a two-stage training regime for a faster convergence. We first pre-train the SDF network using the canonical model mentioned in Sec~\ref{construction_features} and then train all the networks jointly. The learning rate is first linearly warmed up from $0$ to $5\times 10^{-4}$ in the first 2k iterations and then is controlled by the cosine decay schedule. In the second stage, we freeze the SDF network after training for 50K iterations. We use the sampling strategy proposed in NeuS~\cite{wang2021_neus}, and the numbers of the coarse and fine sampling are 32 and 32 respectively. Following Neural Actor ~\cite{liu2021neuralactor}, we also adopt a geometry-guided ray marching process for the volume rendering. Specifically, we only sample points near the SMPL surface to speed up the rendering process. We first train our models on 4 Nvidia V100 32G GPUs and sample 1024 rays per batch per GPU for 80K iterations without LPIPS loss.
Then, we randomly select two additional patches $P$ (size $24 \times 24$) per GPU to continue training the model with LPIPS loss for 20,000 iterations.
The training takes about $38+18$ hours to complete. Because the number of subjects in the ZJU-MoCap is small, we augment the data using color jittering during the training. For the loss weight, we set $\lambda_1, \lambda_2$ both to 0.1, and $\lambda_3$ to 1.

\textbf{Evaluation metric.}
We measure the quality with two evaluation metrics: Peak Signal-to-Noise Ratio (PSNR), and Structural Similarity Index Measure (SSIM). Following previous work, we project the 3D bounding box of the human body onto the image plane to get a 2D mask and only calculate the PSNR and SSIM of the mask area instead of the whole image. For the 3D reconstruction, we only provide the qualitative results, as shown in Fig.~\ref{fig:mesh_rusults}, because the ground truth is not available.

The goal of our work is to get novel-view and novel pose human synthesis. To the best of our knowledge, only MPS-NeRF\cite{gao2022mps} can achieve both at the same time. Hence, we conduct two experiments: novel view synthesis generation (Sec~\ref{gene_exp}) and pose control animation (Sec~\ref{animation_exp}).

\setlength{\tabcolsep}{10pt}
\begin{table*}[tb!]
\renewcommand\arraystretch{1.30}
\begin{center}
\caption{Quantitative comparison of the generalization task in the four settings. We evaluate the synthesis quality on two metrics: PSNR and SSIM . 
We compare with other generalized model, Neural Human Performer (NHP)~\cite{kwon2021neural_gene_human1},Keypoint NeRF (KN)~\cite{kpnerf} and MPS-NeRF (MPS)~\cite{gao2022mps}. Our method achieves significantly better performance in identity  and cross-dataset generalization. In other two setting, our method is on par with other methods. 
\reversion{Additionally, we provide the results of a person-specific model, Neural Body (NB) \cite{peng2021neural_human_nerf_2}, in the task of seen poses for seen subjects as an upper-bound baseline. Our method achieves comparable performance with this benchmark.}
}
\label{table:novelview}
\begin{tabular}{l|ll|ll|ll|ll}
\hline
 Setting& \multicolumn{2}{c|}{Identity generation}&
 \multicolumn{2}{c|}{ Cross-dataset }&
 \multicolumn{2}{c|}{ Pose generation }&
 \multicolumn{2}{c}{ Seen subjects}\\
\hline
Method  & PSNR$\uparrow$& SSIM$\uparrow$ & PSNR$\uparrow$& SSIM$\uparrow$   & PSNR$\uparrow$& SSIM$\uparrow$ &PSNR$\uparrow$& SSIM$\uparrow$\\
\toprule
NB  & - & -  & - & -& -  & - & 28.56 & 0.943\\
\bottomrule
NHP   & 24.85 & 0.908  & 23.16 &  0.869   &26.19& 0.915 &  26.90 & 0.927   \\
KN   & 24.92 & 0.910  &  22.31 & 0.861  &\textbf{27.64}& \textbf{0.933} &  \textbf{27.91} & \textbf{0.938}  \\
MPS   & 22.99 &  0.877  & 22.38 & 0.842  &24.66& 0.880 &  24.84 & 0.887   \\
Ours  &\textbf{25.14} & \textbf{0.914}& \textbf{24.19} & \textbf{0.886} &27.36 & 0.929  &27.83 &\textbf{0.938} \\
\hline
\end{tabular}
\end{center}
\end{table*}

\subsection{Generalization 
\label{gene_exp} 
}
In this part, we evaluate our approach on the novel view synthesis generalization task. 
Given a sparse multi-view (e.g., 3 or 4 views) videos of an unseen person performing arbitrary motions, our method can synthesize novel views of the person performing these motions without training.
In this setting, we compare our method with MPS-NeRF (MPS) \cite{gao2022mps},
Keypoint NeRF (KN) \cite{kpnerf}, Neural Human Performer (NHP) \cite{kwon2021neural_gene_human1} and Neural Body (NB) \cite{peng2021neural_human_nerf_2}.
We also compare our method with Neural Body (NB) \cite{peng2021neural_human_nerf_2} which is a person-specific model

ZJU-MoCap dataset consists of  $10$ human subjects captured from 23 synchronized cameras.
Following NHP, we split the dataset into two parts: $7$ \textbf{source}  subjects and $3$ \textbf{target}  subjects.
We evaluate our method and the baseline methods in the following four different settings.
Note that for all the comparisons except the cross-dataset generalization, the first $300$ frames of the source or target subjects are used for training, and the rest frames (unseen poses) are used for testing.

\textbf{1) Identity generalization.}
First, we evaluate the generalization to different identities by testing on the \textbf{target subjects}.
\reversion{All methods are trained on all the source subjects and directly tested on target subjects. As Tab. \ref{table:novelview} and Fig. \ref{fig:unseen_people} show, our method gives the best performances quantitatively and qualitatively.} 

\textbf{2) Cross-dataset generalization.}
To further test the generalizability of our method to new datasets, we train all methods on the ZJU-MoCap dataset  and directly test on the DeepCap~\cite{habermann2020deepcap}  and DynaCap~\cite{habermann2021real_DynaCap} dataset  without fine-tuning. As shown in the Tab. \ref{table:novelview}, our method significantly improves the performance compared to other methods.
Even though the training and testing datasets are  significantly different in the appearance distribution and the distance from the camera to subject, our method still can achieve impressive results without fine-tuning, as shown in Fig.~\ref{fig:cross_dataset}.


\textbf{3) Pose generalization.}
In this setting, all the methods are trained on the source subjects and tested on \textbf{unseen poses of the same subject}. \reversion{ All methods are trained on all source subjects together.
As Tab. \ref{table:novelview} shows, our model outperforms MPS, NHP significantly on PSNR and SSMI. Our method also achieve comparable results to KN. }

\textbf{4) Seen poses of seen subjects.}
To demonstrate the superiority of our model, we evaluate the performance for seen poses of source subjects. All methods except NB are trained on the all source subjects and tested on the seen poses of them.
Tab. \ref{table:novelview} demonstrates that our method outperforms MPS and NHP, and it is comparable to NB and KN.

Note that KN is sightly better than our method in task $3$ and $4$, but our method outperform it significantly in task $1$ and $2$.
Moreover, there are artifacts in the results of KN for cross dataset, which do not appear in the results of the ZJU-MoCap dataset.
The reason for this is that KN sets a fixed hyper-parameter that controls the impact of each keypoint. The hyper-parameter determines relative spatial encoding for 3D query point and keypoints, which is sensitive to human body shape and pose. As a result, KN can overfit the training dataset, but it does not generalize well for the pose or shape not seen in the training dataset. And KN is not animatable human models.

Our method shares the same goal as MPS-NeRF but differs in our approach. MPS-NeRF mainly relies on the SMPL model to fuse image features for generating novel views, while we associate surface features with the SMPL model and diffuse them onto surrounding spatial points using a specially designed GCN network. This allows us to reconstruct reasonable geometry details, such as clothes and hair, even with sparse input images. Additionally, our occlusion-aware self-attention mechanism can effectively handle mesh points that are occluded in the sparse input images. As shown in Tab. \ref{table:novelview}, these mechanisms allow our method to significantly outperform MPS-NeRF in all tasks.

In conclusion, our method achieves state-of-the-art results on the novel view synthesis generalization task for new character (identity and cross-dataset generalization). And our method is comparable to other methods in the fitted task. 



\begin{figure*}[t]
    \centering
    \includegraphics[width=0.95\linewidth]{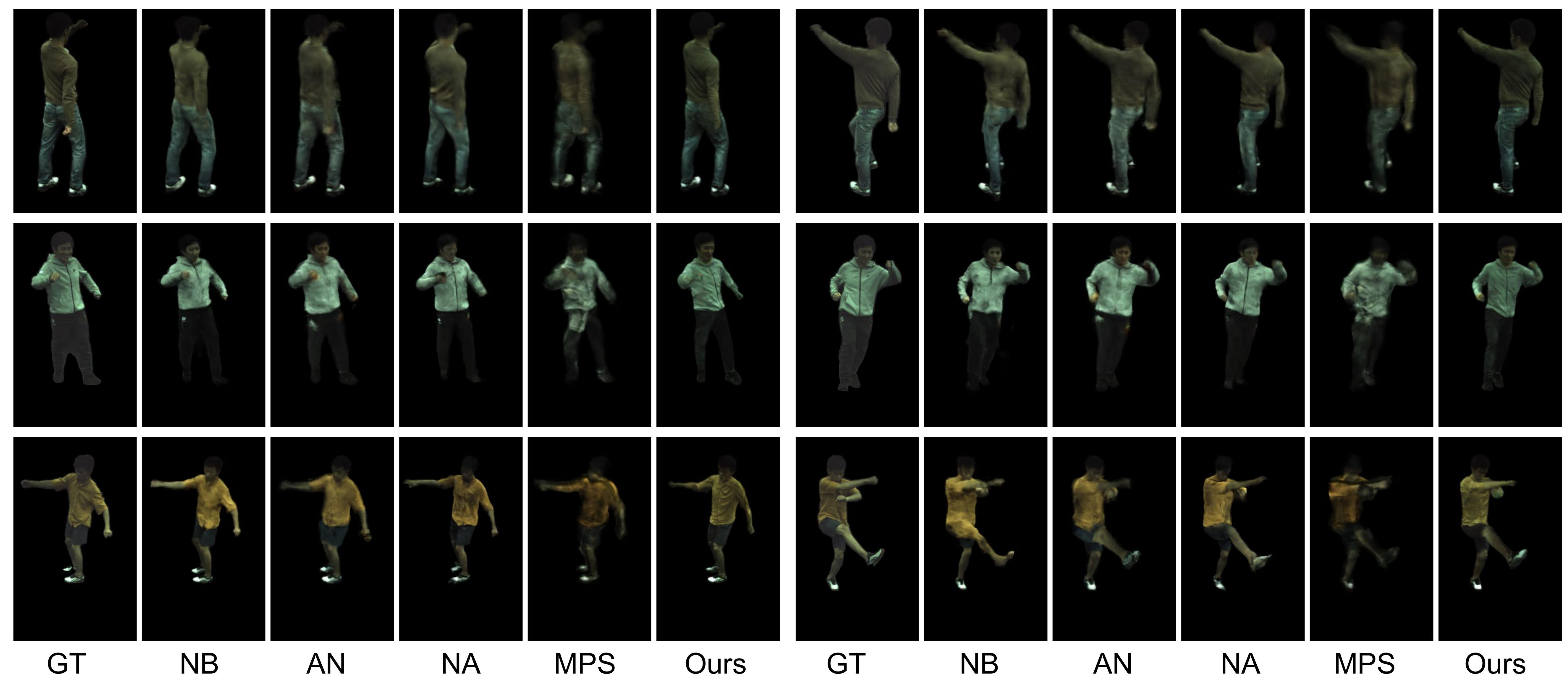}
    \caption{Qualitative comparison of the Animation task on the ZJU-MoCap dataset. We compared with three person-specific methods, Neural Body (NB)\cite{peng2021neural_human_nerf_2}, Animatable Nerf (AN) \cite{peng2021animatable}, Neural Actor (NA)\cite{liu2021neuralactor} and  the generalized method, MPS-NeRF (MPS) \cite{gao2022mps}. Note that the person-specific methods are trained with $300$ frames of a target person and tested on the same person, while MPS-NeRF and our method is trained on all source subjects and obtains an animatable human model of a target person just from \textbf{one} frame of the person. Our method outperforms all baselines even in a disadvantaged setting.
    }\label{fig:animation}
\end{figure*}

\begin{figure*}[t]
    \centering
    \includegraphics[width=0.95\linewidth]{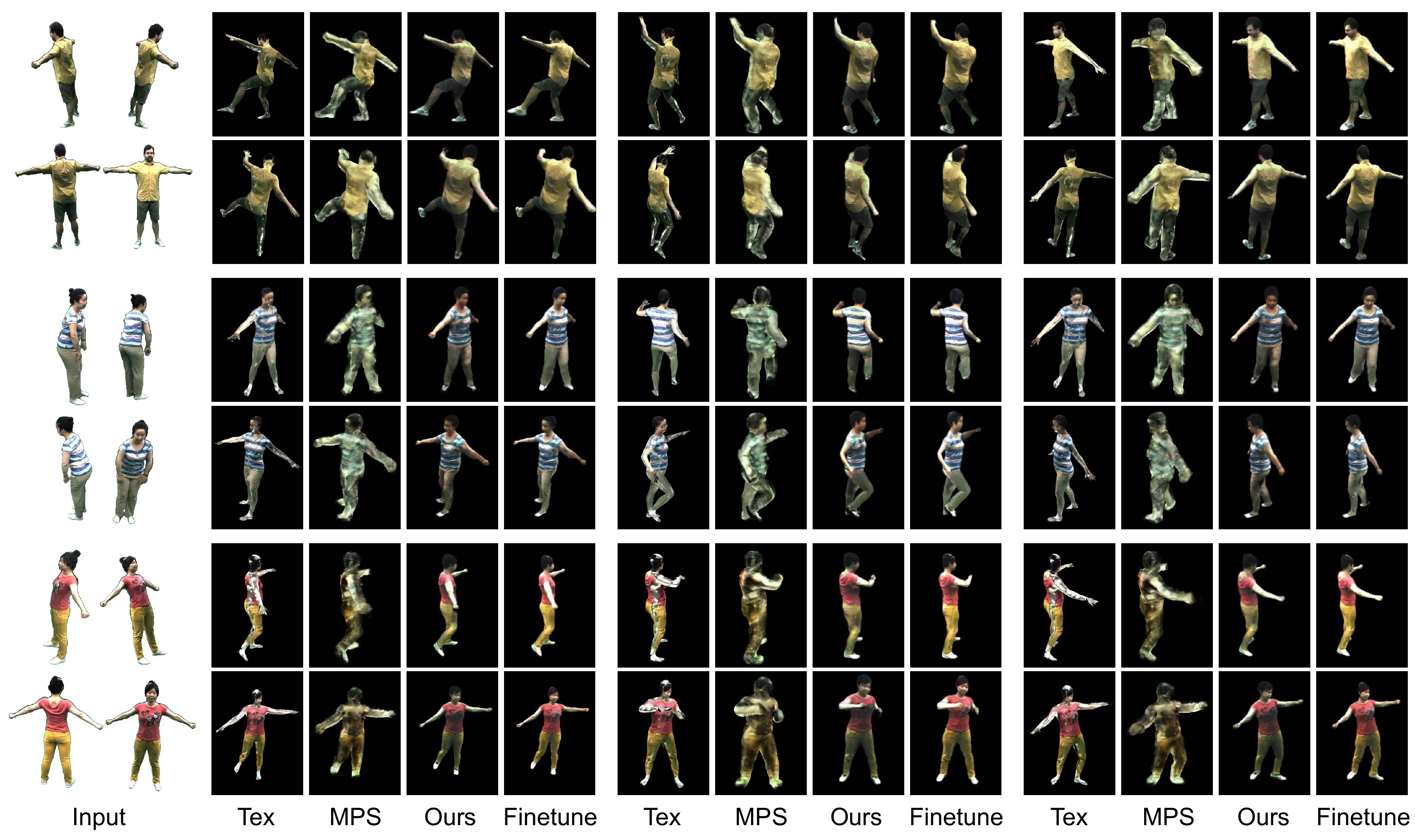}
    \caption{\reversion{Qualitative comparison of cross-dataset generalization on animation. Tex:  projects pixel colors onto the surface of the SMPL model with per-vertex deformations. Finetune: fine-tune our model for approximately 10 minutes with the same input. MPS-NeRF and our method trained on the ZJU-MoCap~\cite{peng2021neural_human_nerf_2}  and directly tested on the DeepCap~\cite{habermann2020deepcap} and DynaCap~\cite{habermann2021real_DynaCap}. Our method shows significant superiority over other methods, while also yielding superior outcomes with fine-tuning.
    }}\label{fig:cross_posedriven}
\end{figure*}

\subsection{Animation  \label{animation_exp}}
The task of animation is that, given sparse multi-view images of an \textbf{unseen} person in a static pose, the model needs to  generate novel pose images under user’s pose control.
\reversion{This task is distinct from the pose generation task discussed in Sec~\ref{gene_exp}. In the case of generating images, the task only requires pose input without any image input. On the other hand, pose generation tasks involve using image input to generate novel view images.}
To evaluate the performance of this task, we compare our method with Neural Body (NB), Animatable Nerf (AN) \cite{peng2021animatable},  Neural Actor (NA) \cite{liu2021neuralactor} and MPS-NeRF (MPS) \cite{gao2022mps} on the ZJU-MoCap dataset. Neural Human Performer (NHP) \cite{kwon2021neural_gene_human1} and Keypoint NeRF (KN) \cite{kpnerf} are not animatable human models and thus cannot generate images with user’s pose control. The spilt for the ZJU-MoCap dataset is the same as that described in Sec. \ref{gene_exp}.

MPS-NeRF and our method are the generalized animatable human model for this task, so we train them on all the source subjects. At test time, we directly obtain an animatable model of the target person just from the sparse camera views of one frame of the target person without training. 
Other person-specific methods are trained in a person-specific manner on the first $300$ frames of the target person and tested on the remaining frames of the same person.

Tab. \ref{table:animation} and Fig. \ref{fig:animation}  demonstrate that our method significantly outperforms MPS-NeRF quantitatively and qualitatively.
MPS-NeRF relies on the inverse skinning over the SMPL model to animate the character, while our method employs an additional residual deformation mapping to compensate for the deformation that SMPL cannot model. This mechanism results in a more accurate shape and appearance, as shown in Fig. \ref{fig:animation}. 

Moreover, our method outperforms other baselines even though it is trained and evaluated in a more difficult setting (i.e., unlike other baselines, our method is not overfitted to the target subject during training). \reversion{Neural Actor (NA) \cite{liu2021neuralactor} is designed for input videos with dense camera views and requires obtaining textures for each frame of the input video. However, in sparse view settings, acquiring high-quality textures can be challenging due to issues like self-occlusion that reduce effectiveness. Moreover, Animatable NeRF (AN) \cite{peng2021animatable} and Neural Body (NB) \cite{peng2021neural_human_nerf_2} do not account for self-occlusion in sparse view and lack specific designs for completing the missing parts during training.  In contrast, our method can obtain high-quality avatars from sparse view inputs using carefully designed modules, such as GCN and the occlusion-aware self-attention mechanism. We believe that our design can also enhance per-scene optimization methods. Additionally, our method is trained on multi-person data, which allows for better pose generalization compared to methods trained on single-person data.}

\setlength{\tabcolsep}{9pt}
\begin{table}[t]
\begin{center}
\medskip
\caption{Quantitative comparison of the animation task.  Our method achieves the best performance in two metrics, compared to three person-specific animatable human models,  NeuralBody (NB)~\cite{peng2021neural_human_nerf_2},  AnimatbleNerf (AN)~\cite{peng2021animatable}, and Neural Actor (NA)~\cite{liu2021neuralactor} and MPS-NeRF (MPS) \cite{gao2022mps}}. 
\label{table:animation}
\scalebox{1.0}{\begin{tabular}{c|ccccc}
\hline
Method & NB & AN& NA & MPS &Ours\\
\hline
PSNR$\uparrow$ &23.03 &22.95 &22.50  &21.80&\textbf{23.18} \\ 
SSIM$\uparrow$ &0.880 &0.875 &0.878  &0.858&\textbf{0.886} \\ 
\hline
\end{tabular}
}
\end{center}
\end{table}

\begin{figure}[t]
    \centering
    \includegraphics[width=0.95\linewidth]{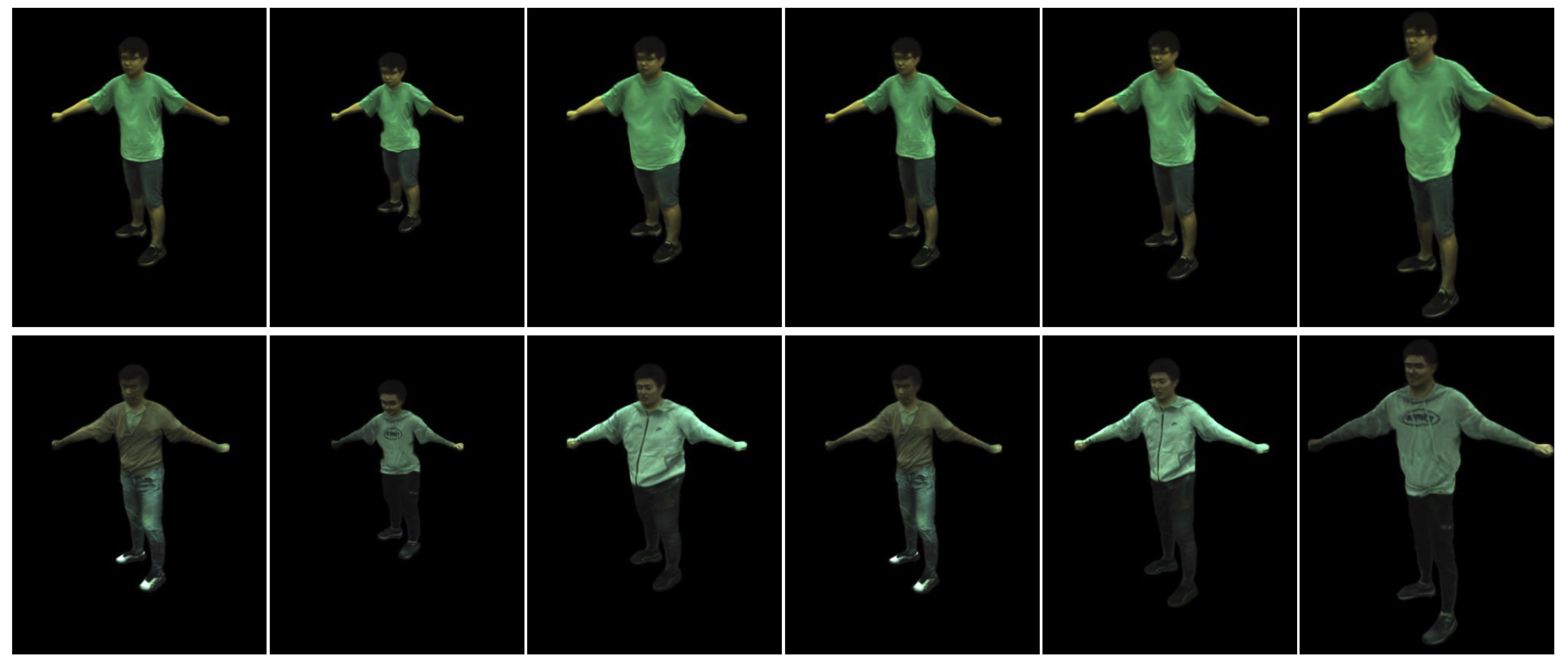}
    \caption{Results of changing appearance and geometry. We can directly change human shape (top row) and human appearance (bottom row) while keeping other factors fixed.}\label{fig:exchange_geo}
\end{figure}

\reversion{To more effectively demonstrate the efficacy of our approach, we assess its performance on a cross-dataset evaluation. 
Among existing methods, only MPS-NeRF and our method are capable of accomplishing this task. 
We train  all method on the ZJU-MoCap dataset and directly test on the DeepCap~\cite{habermann2020deepcap} and DynaCap~\cite{habermann2021real_DynaCap} dataset.
To demonstrate the difficulty of this task, building upon prior research by VideoAvatar~\cite{alldieck2018video_sm_hair_1}, we implemented a baseline projection method (denoted as Tex), which projects pixel colors onto the surface of the SMPL model. The SMPL model, learned from scans of unclothed humans, 	is unable to represent clothing or other personal surface details. We estimate per-vertex deformations of the SMPL model by rendering it to fit the foreground mask. However, it cannot accurately model geometry, due to the limitations of the SMPL template as a base.
In contrast to simple projection, we reconstructed better geometry and used our GCN module to complete the missing parts from a few viewpoints, resulting in better performance, as shown in Fig.~\ref{fig:cross_posedriven}. Furthermore, the state-of-the-art method MPS-NeRF, which relies exclusively on pixel-aligned features and does not incorporate human prior information or disentangle appearance and geometry, produced inferior results, particularly when applied across different datasets. }

\reversion{
We also fine-tune our model for approximately $10$ minutes with the same input, using $3$ images as input and another image as the target each time. As shown in Fig.~\ref{fig:cross_posedriven}, fine-tuning can reduce color deviations caused by differences in lighting distribution between the test data and training dataset. Additionally, fine-tuning can result in more accurate geometry. 
}

\subsection{Appearance and geometry control}
As our method disentangles the geometry and appearance in the human modeling, we can either change the appearance while keeping the geometry fixed or control the body shape of modeled humans by manipulating shape parameters in SMPL. Fig. \ref{fig:exchange_geo} demonstrates the synthesized images after changing the body shape and exchanging the appearance.

\subsection{Ablation Study}
\begin{figure}[t]
    \centering
    \includegraphics[width=0.95\linewidth]{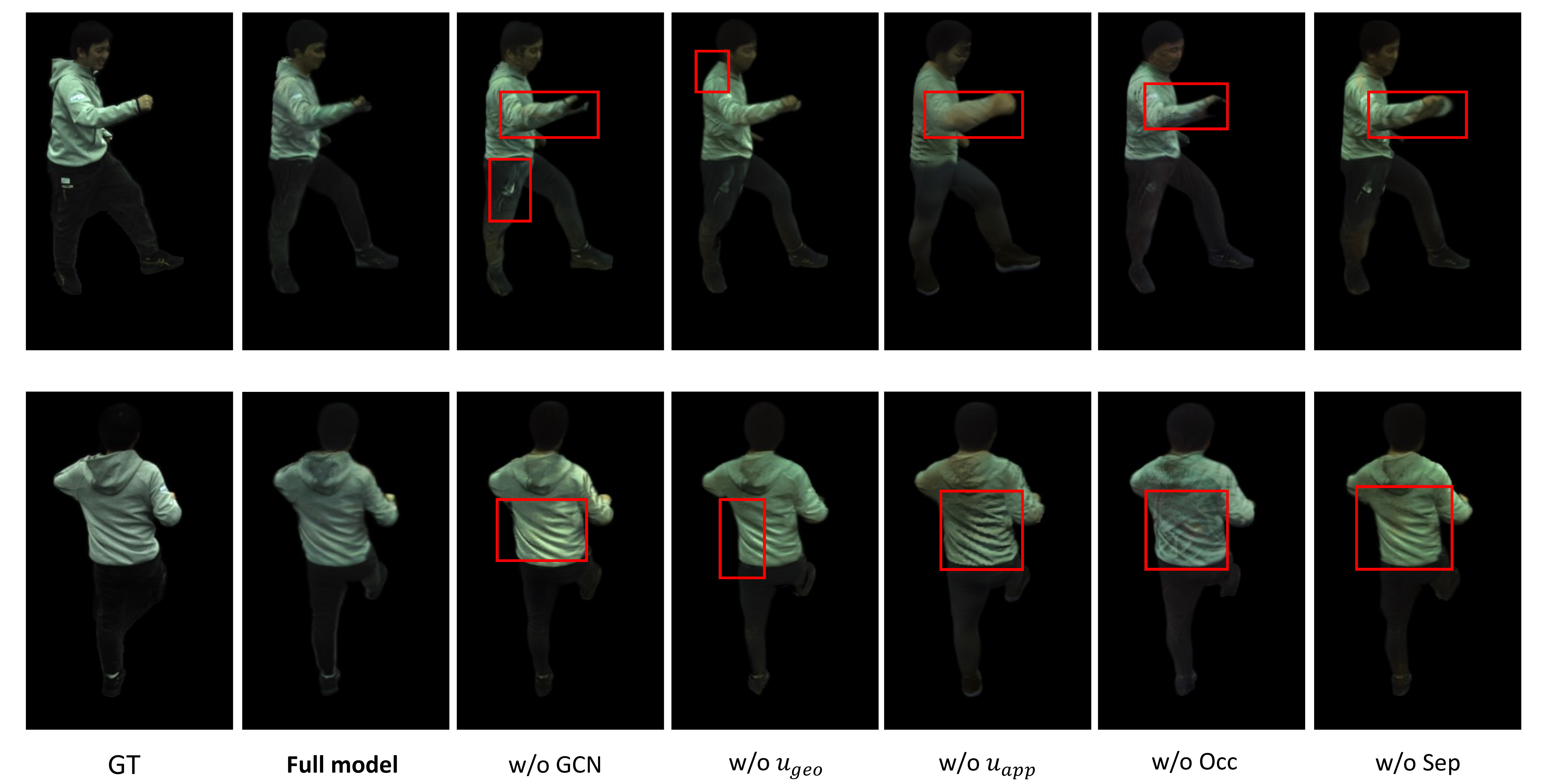}
    \caption{\reversion{Visual result of the ablation studies for different components.
    Occ: the occlusion-aware self-attention mechanism, Sep: separation of geometry and appearance features. 
    }
    }\label{fig:ablation}
\end{figure}

We conduct ablation studies using the ZJU-MoCap dataset on both the generalization and animation tasks. The same experiment settings as described in Sec. \ref{gene_exp} and Sec. \ref{animation_exp} are used for these two tasks. The results are shown in Table \ref{table:ablation}. \reversion{We also provide visual result in Fig.~\ref{fig:ablation} to further demonstrate the significance of different components.}

We first evaluate the effect of the GCN used in extracting the surface features. The baseline, w/o GCN, is performed by computing the surface features directly using the image features. \reversion{In the absence of the GCN module, the model cannot effectively handle occlusion and utilize prior knowledge about the human body as discussed in Sec.~\ref{skel_feature}, which leads to artifacts both in shape and appearance as showed in Fig.~\ref{fig:ablation}. Additionally, the drop in performance shown in Tab.~\ref{table:ablation} underscores the importance of the GCN module.}

We also evaluate the effect of the image features. We compare with: 1) removing the image features from the geometry features (w/o $\vect{u}_{\text{geo}}$); 2) removing the image features from the appearance features (w/o $\vect{u}_{\text{app}}$); 3) disabling the occlusion-aware self-attention mechanism (w/o Occ) by letting $B^c = 0$ in \eqnref{\eqref{eqn:occ_bias_att}}.
\reversion{As visible in Fig.~\ref{fig:ablation}, the predicted results have significant geometric deviations when lacking the geometry feature $\vect{u}_{\text{geo}}$, mainly reflected in the lack of clothing geometry and errors in the shape of the person. Similarly, when lacking the appearance feature $\vect{u}_{\text{app}}$, the model's generalization ability is significantly impaired, as evidenced by reconstructing incorrect clothing colors and confusing clothing with limbs. Furthermore, when lacking the occlusion module, the model's ability to process multi-view information is greatly reduced, and it may incorrectly utilize the input information that should not be used, such as reconstructing frontal clothing information in the back.}
These comparisons show that the image features are critical to quality results, while our occlusion-aware mechanism effectively improves the perceptual realism.

To validate our design of disentangling the geometry and appearance features, we train a model with a single variable for both the appearance and geometry features (w/o Sep). \reversion{When the disentangling module is missing, the model is prone to confusing geometry and appearance during training, resulting in many artifacts in both geometry and appearance in the predicted results. For example, in Figure~\ref{fig:ablation}, an erroneous human body shape is depicted, and clothing is mistakenly reconstructed on the hand.} As shown in Tab. \ref{table:ablation}, our disentangled features achieve better image quality in terms of all of the three metrics. 

In table \ref{table:ablation_view}, we show the performance of our model in terms of the generalization to novel views and novel animations with different numbers of input views. The performance of our method degrades slightly when given fewer input views.

\begin{figure}[t]
    \centering
    \includegraphics[width=0.95\linewidth]{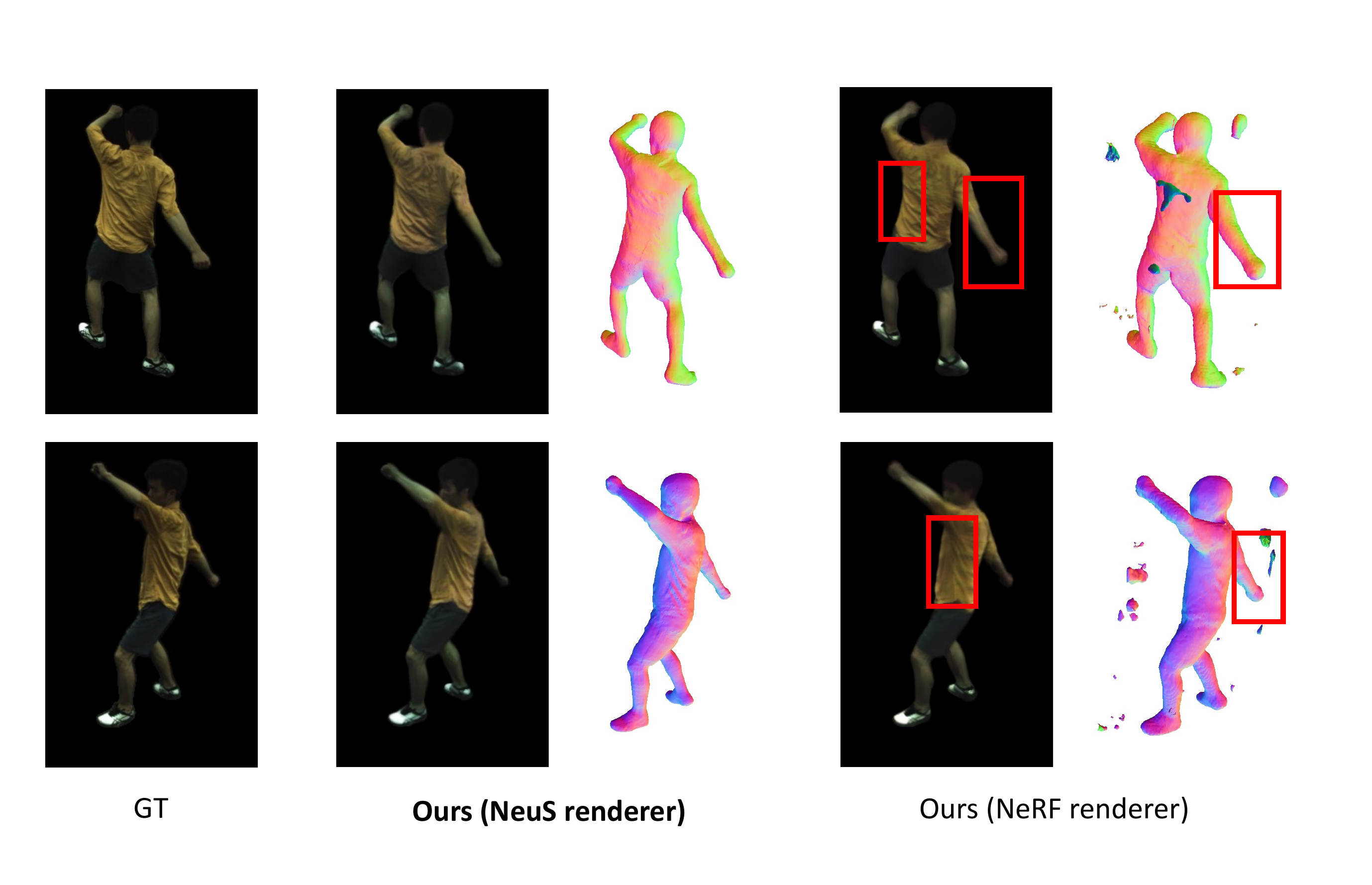}
    \caption{\reversion{ Comparison of NeuS rendering and NeRF rendering: The NeuS rendering method enables more accurate human geometry and fewer appearance artifacts.
    }
    }\label{fig:neus_vs_nerf}
\end{figure}

We also compared our method using NeuS and the original volume rendering algorithm in Nerf. By using the NeuS rendering method, the sampling points are more concentrated on the object surface, which makes the deformation field easier to train. This also allows for more accurate human geometry and fewer appearance artifacts, as shown in the pose driven results in the Fig.~\ref{fig:neus_vs_nerf}.

\setlength{\tabcolsep}{10pt}  
\begin{table}[t]
\renewcommand\arraystretch{1.3} 
\begin{center}
\caption{Ablation study for different components.
Occ: the occlusion-aware self-attention mechanism, Sep: separation of geometry and appearance features. 
}
\label{table:ablation}
\scalebox{0.88}{\begin{tabular}{c|cc|cc}
\hline
& \multicolumn{2}{c|}{Unseen subjects}& \multicolumn{2}{c}{Animation}\\
\hline
& PSNR$\uparrow$ & SSIM$\uparrow$ &  PSNR$\uparrow$ & SSIM$\uparrow$\\
\hline
w/o GCN & 24.27 &  0.891  & 22.81 & 0.872 \\ 
w/o $\vect{u}_{\text{geo}}$ & 23.79 & 0.890   &22.34 & 0.875  \\
w/o $\vect{u}_{\text{app}}$  & 23.39 & 0.881  & 22.38 & 0.870 \\ 
w/o Occ & 24.10 & 0.889 & 22.79 & 0.874 \\
w/o Sep & 23.68 & 0.891 & 22.46 &0.874   \\
\hline
Full model & \textbf{25.14} &\textbf{0.914}  &   \textbf{23.18} & \textbf{0.886}\\  
\hline
\end{tabular}
}
\end{center}
\end{table}

\setlength{\tabcolsep}{2.5pt}
\begin{table}[t]
\begin{center}
\caption{Quantitative evaluation of using different numbers of input camera views during inference. The performance of our method degrades slightly when given fewer input views.} 
\label{table:ablation_view}
\begin{tabular}{cc|ccccc}
\hline
 view number& & & 1 view & 2 views & 3 views & 4 views\\ 
\hline
\multicolumn{1}{c}{\multirow{3}{*}{Unseen subjects }} &PSNR$\uparrow$ && 23.28 &24.26 & 24.78 & \textbf{25.14}\\
\multicolumn{1}{c}{} & SSIM$\uparrow$ &&0.887 & 0.900& 0.909 &\textbf{0.914}\\
\hline
\multicolumn{1}{c}{\multirow{3}{*}{Animation}} &PSNR$\uparrow$ && 22.30 & 22.90 &23.08 &\textbf{23.18} \\
\multicolumn{1}{c}{} & SSIM$\uparrow$ && 0.875 & 0.881 & 0.883 & \textbf{0.886} \\ 
\hline
\end{tabular}
\end{center}
\end{table}

\section{Discussion}
Despite its success, our framework still has several limitations.
First, our method relies on the accuracy of pose tracking, and the low-quality SMPL estimation may result in artifacts. It would be interesting to optimize the SMPL parameters in the framework.
Second, our method can only handle the clothing types that follow the topology of the SMPL model, and it is challenging to model very loose clothes, such as skirts.
Finally, our method does not model complex lighting effects and we assume the uniform lighting in the input multi-view videos. When the assumption cannot be met (e.g.,  the ZJU-MoCap dataset), our model tends to learn the average lighting and produce the results with color shift.\reversion{ We leave them for future work. Moreover, it is noteworthy that our proposed model exhibited exceptional performance despite being trained on a restricted dataset. The potential benefits of augmenting our framework to accommodate larger models and incorporating additional training data warrant further exploration.
}

\reversion{We extract image features from both the spatial and surface points of the SMPL model to more effectively infer person- and pose-dependent properties, and our GCN integrates local pose information by using the edges of the posed SMPL as its edge features, enabling the generalization of pose-aware surface features. 
In contrast, for the novel view synthesis generalization task, the state-of-the-art method, Keypoint NeRF (KN)~\cite{kpnerf}, relies on spatial encoding for 3D query points and keypoints, which is sensitive to human body shape and pose. 
The person- and pose-dependent properties of our method lead to improved performance in pose-dependent appearance and geometry in the task of cross-dataset and identity generalization. On the other hand, in the task of animation, although  we design a residual module to compensate for the deformation caused by changes in body shape and pose, using the extracted pose-aware surface features. However, determining the changes in appearance and geometry with different poses is challenging, given only four static images as input. Consequently, the performance of pose-dependent appearance and geometry in this task is not as strong as in novel view synthesis.
As a result, exploring the use of multi-view videos in testing to enhance this capability is another interesting direction.
}
\section{Conclusion}
We presented Neural Novel Actor, a new method for learning a generalized animatable neural human representation from a sparse set of multi-view imagery of multiple persons.
With the learned representation, we can synthesize novel view images of an arbitrary person using a sparse set of cameras and further synthesize animations with user’s pose control. 
To efficiently learn this representation for multiple persons, we design our proposed human representation with disentangled geometry and appearance. Furthermore, we leverage the features at both the spatial points and the surface points of SMPL to infer pose- and person-dependent geometry and appearance. 
Extensive experiments demonstrate that our method significantly outperforms the state-of-the-arts on the tasks of novel view synthesis of new persons and the animation synthesis with pose control. 

\ifCLASSOPTIONcompsoc
  \section*{Acknowledgments}
\else
  \section*{Acknowledgment}
\fi

This work was supported in part by NSFC Projects of International Cooperation and Exchanges (62161146002). Christian Theobalt was supported by ERC Consolidator Grant 4DReply  (770784). Lingjie Liu was supported by Lise Meitner Postdoctoral Fellowship.

\bibliographystyle{abbrv-doi-hyperref}
\bibliography{egbib}

\ifCLASSOPTIONcaptionsoff
  \newpage
\fi

\begin{IEEEbiography}[{\includegraphics[width=1in,height=1.25in,clip,keepaspectratio]{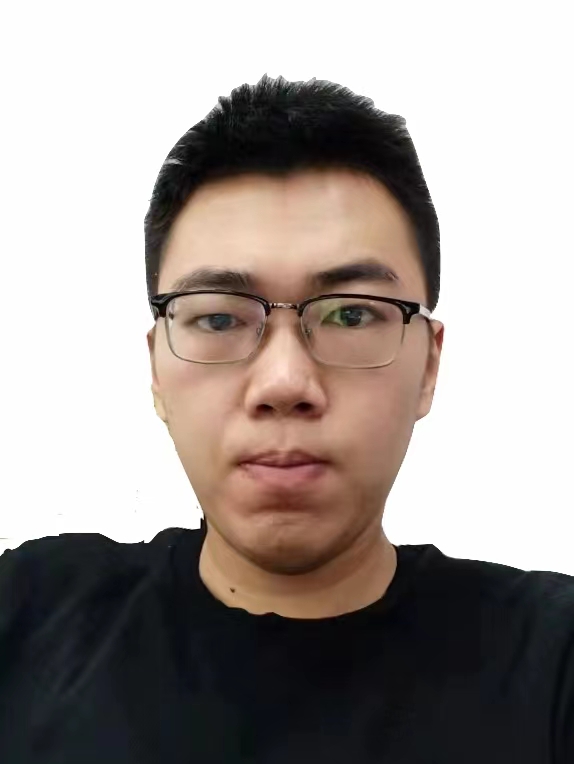}}]{Qingzhe Gao} is a a forth-year Ph.D. student in
the School of Computer Science and Technology, Shandong University. His research interests include e 3D reconstruction, image segmentation, and neural rendering.
\end{IEEEbiography}
\begin{IEEEbiography}[{\includegraphics[width=1in,height=1.25in,clip,keepaspectratio]{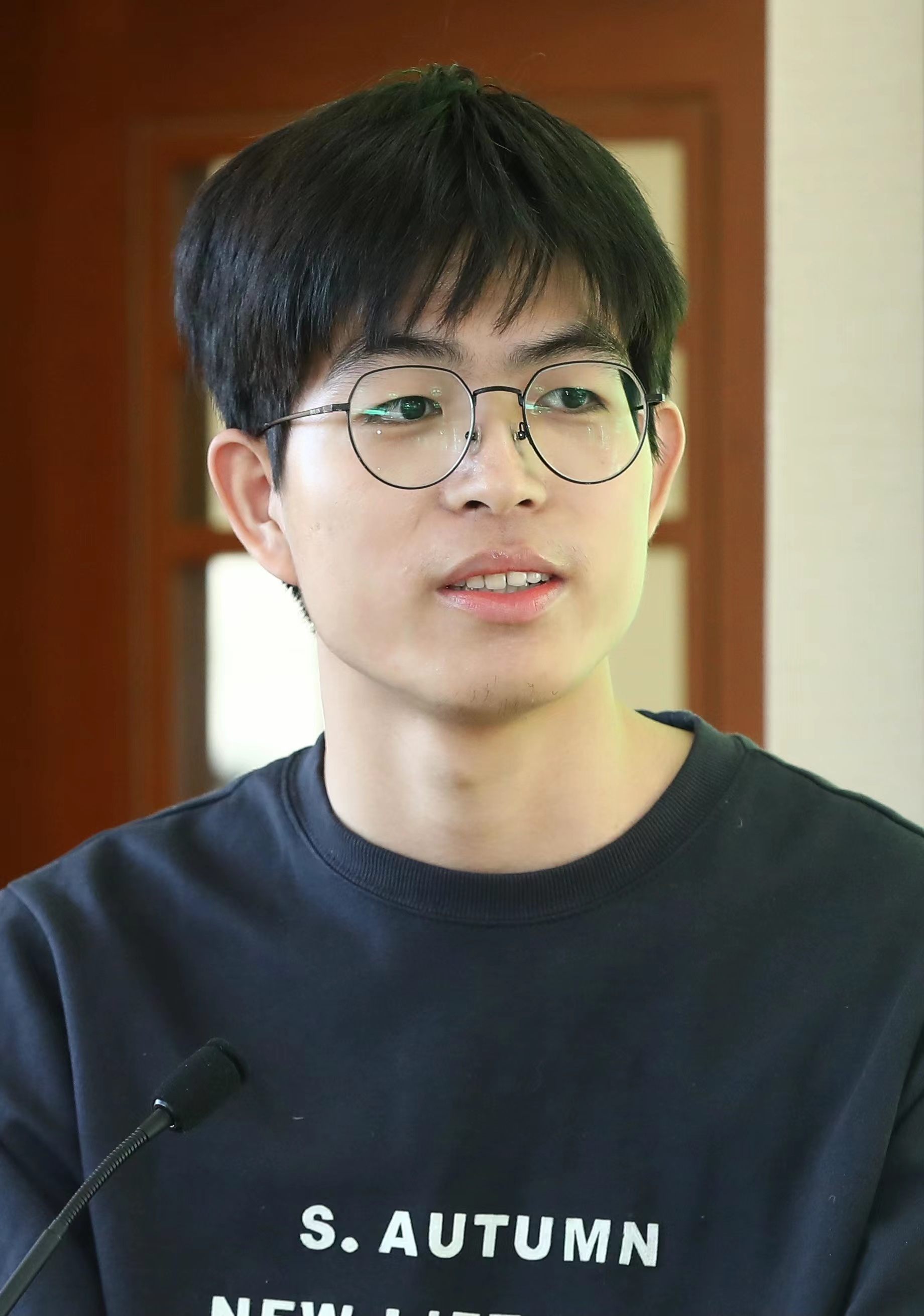}}]{Yiming Wang} is a four-year undergraduate student in the School of Electronics Engineering and Computer Science, Peking University. His research interests include neural scene representations and 3D reconstruction.
\end{IEEEbiography}
\begin{IEEEbiography}[{\includegraphics[width=1in,height=1.25in,clip,keepaspectratio]{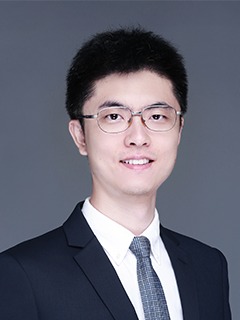}}]{Libin Liu} is an assistant professor at the School of Intelligence Science and Technology, Peking University. He received his Ph.D. degree in computer science from Tsinghua University. His research interests include character animation, physics-based simulation, motion control, and related areas in machine learning and robotics.
\end{IEEEbiography}
\begin{IEEEbiography}[{\includegraphics[width=1in,height=1.25in,clip,keepaspectratio]{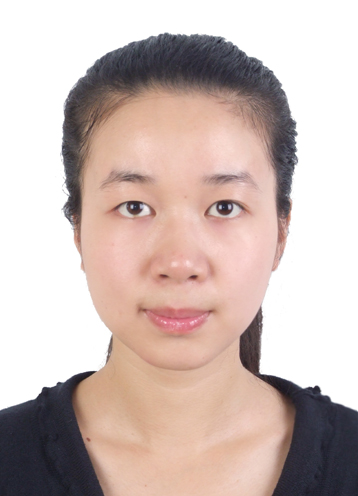}}]{Lingjie Liu} is a post-doctoral researcher at the Graphic, Vision \& Video group of Max Planck Institute for Informatics in Saarbrücken, Germany. She received her BEng degree from the Huazhong University of Science and Technology in 2014 and PhD degree from the University of Hong Kong in 2019. Her research interests include 3D reconstruction, neural rendering and human performance capture. She has received Hong Kong PhD Fellowship Award (2014) and Lise Meitner Postdoctoral Fellowship Award (2019).
\end{IEEEbiography}
\begin{IEEEbiography}[{\includegraphics[width=1in,height=1.25in,clip,keepaspectratio]{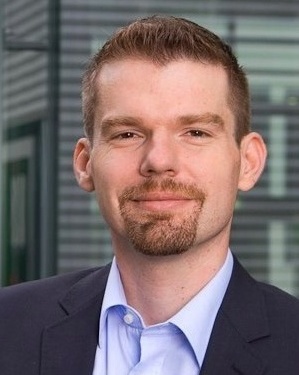}}]{Christian Theobalt} is a Professor of Computer Science and the head of the research group “Graphics, Vision, \& Video” at the Max-Planck-Institute for Informatics, Saarbruecken, Germany. He is also a professor at Saarland University. His research lies on the boundary between Computer Vision and Computer Graphics. For instance, he works on 4D scene reconstruction, marker-less motion and performance capture, machine learning for graphics and vision, and new sensors for 3D acquisition. Christian received several awards, for instance the Otto Hahn Medal of the Max-Planck Society (2007), the EUROGRAPHICS Young Researcher Award (2009), the German Pattern Recognition Award (2012), an ERC Starting Grant (2013), and an ERC Consolidator Grant (2017). In 2015, he was elected one of Germany’s top 40 innovators under 40 by the magazine Capital. He is a co-founder of theCaptury (www.thecaptury.com).\end{IEEEbiography}
\begin{IEEEbiography}[{\includegraphics[width=1in,height=1.25in,clip,keepaspectratio]{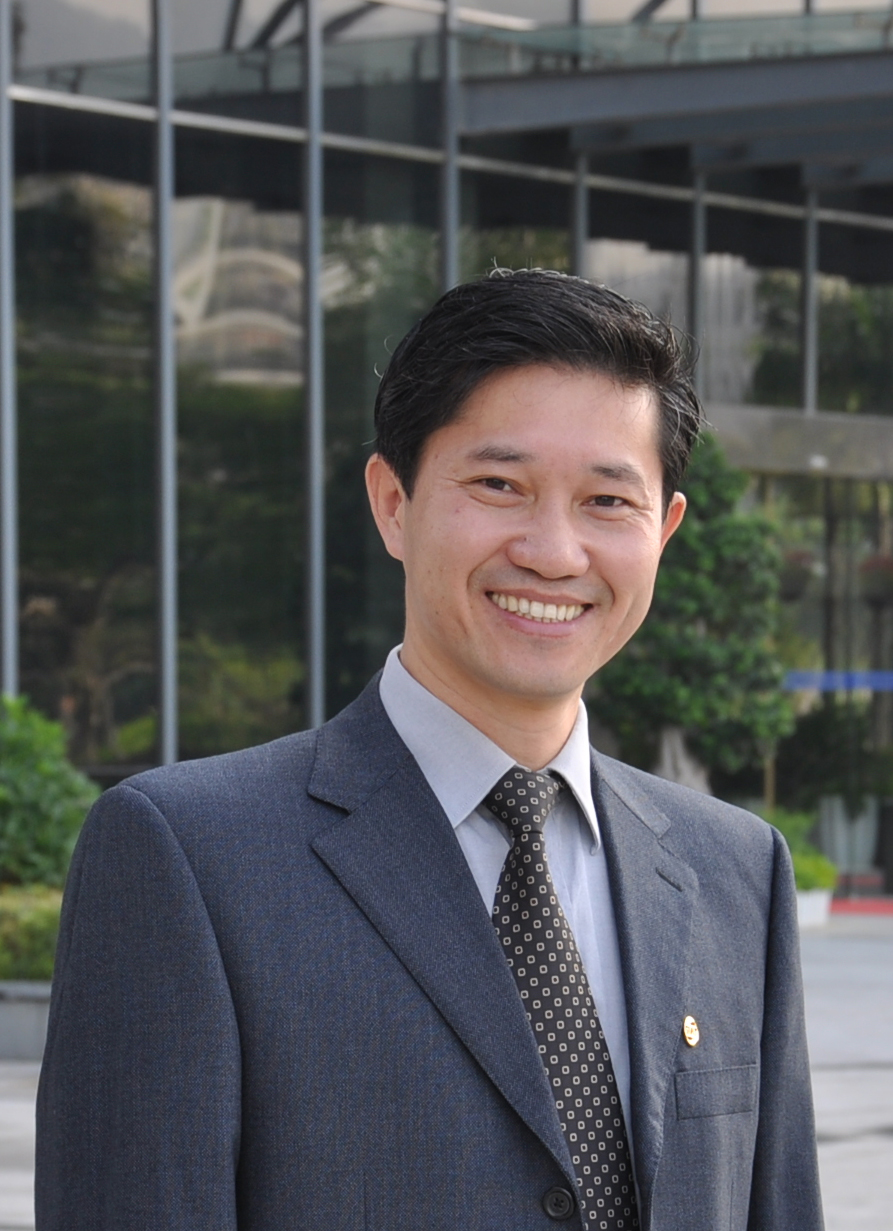}}]{Baoquan Chen} is a professor of Peking University with research interests in computer graphics, visualization and computer vision. He has published more than 200 papers and has received the Best Paper Awards for IEEE Visualization 2005 and ACM SIGGRAPH Asia 2022 and honorary mention for ACM SIGGRAPH 2022. He was elected IEEE Fellow in 2020 and was inducted to IEEE Visualization Academy in 2021. He served as conference chairs of ACM SIGGRAPH Asia 2014 and IEEE Visualization 2005, associate editor of ACM Transactions on Graphics and IEEE Transactions on Visualization and Graphics, and many other roles in professional organizations. 
\end{IEEEbiography}

\end{document}


\maketitle
\section{Dataset}
Following Neural Human Performer~\cite{kwon2021neural_gene_human1}, we split the ZJU-MoCap~\cite{peng2021neural_human_nerf_2} dataset into source and target parts and use them for training and testing respectively.  We also use the DeepCap~\cite{habermann2020deepcap} and DynaCap~\cite{habermann2021real_DynaCap} datasets to evaluate our methods and Neural Human Performer~\cite{kwon2021neural_gene_human1} in terms of cross-dataset generalization. None of these datasets contains personally identifiable information or offensive content.

\subsection{ZJU-MoCap}
The ZJU-MoCap~\cite{peng2021neural_human_nerf_2} dataset provides $10$ dynamic human videos captured using a multi-camera system. It uses EasyMocap\footnote{https://github.com/zju3dv/EasyMocap} to estimate SMPL parameters from multi-view videos and extracts foreground masks using PGN~\cite{gong2018PGN}.  
In our test, we reversed $7$ subjects (index 313, 315, 377, 386, 390, 392, 396)  as the \textbf{source} subjects for the training and $3$ subjects (index 387, 393, 394) as the \textbf{target} subjects for testing on unseen identities.
We use four camera views as the input to test the trained models and evaluate the novel-view images synthesized for the remaining views.  We test our method and the other baselines on the images sampled every 30 frames from the videos, which are resized to $512\times{}512$.


\subsection{DeepCap and DynaCap}
We use two sequences, S1 and S2, from the DeepCap dataset~\cite{habermann2020deepcap}, which contain 11 and 12 cameras respectively.  Both sequences contain videos with around 20,000 frames at the resolution of $1024\times1024$.  We also use another sequence, D1, from the DynaCap dataset~\cite{habermann2021real_DynaCap}.  This sequence is captured with 50 cameras at the resolution of $1285\times940$ and has around 7, 000 frames.  
We estimate the SMPL parameters for these two datasets using the EasyMocap framework and use color keying to extract the foreground masks in each image.  
Similar to the tests on the ZJU-MoCap dataset, we use four camera views as the input to the models and the remaining cameras for comparison.
We test Neural Human Performer~\cite{kwon2021neural_gene_human1} and our model on the same set of images sampled every $100$ frames from these three sequences. These images are then resized to $512\times512$ before feeding to the models.
\section{Additional Result}

\textbf{Video Results.}
We provide a supplementary video for the readers to better measure the visual quality of the results. We encourage the readers to check our video. 

\textbf{Reliability of the results.}
We train 5 models with different random seeds and observe standard deviations ranging from $0.09$ PSNR to $0.16$ PSNR in all experiment settings.

\setlength{\tabcolsep}{13pt}
\begin{table*}[t]
\begin{center}
\caption{Task comparison of our method with the state-of-the-art human rendering methods: 
NeuralBody (NB) \cite{peng2021neural_human_nerf_2},  Animatable NeRF (NA) \cite{peng2021animatable}, Neural Actor \cite{liu2021neuralactor}, Neural Human Performer (NHP) \cite{kwon2021neural_gene_human1}, HumanNeRF$_1$ (HN$_1$) \cite{zhao2021humannerf_gene_human2}, HumanNeRF$_2$ (HN$_2$) \cite{singHumanNeRF}, KeypointNeRF (KN) \cite{kpnerf}, MPS-NeRF (MPS) \cite{gao2022mps}. 
MPS-NeRF and our method are for both the generalization of view synthesis and animation tasks. 
}
\label{table:method}
\begin{tabular}{lllllllllll}
\hline
Task &Setting& NB &AN &NA& NHP&HN$_1$&HN$_2$ &KN&MPS&Ours\\
\hline
Novel view  &seen pose               &\ding{51} &\ding{51} &\ding{51} &\ding{51} &\ding{51}&\ding{51}&\ding{51}&\ding{51} &\ding{51}\\ 
            &unseen pose             &\ding{51} &\ding{51} &\ding{51} &\ding{51} &\ding{51}&\ding{55}&\ding{51}&\ding{51} &\ding{51}\\ 
            &unseen person           &\ding{55} &\ding{55} &\ding{55} &\ding{51} &\ding{51}&\ding{55}&\ding{51}&\ding{51} &\ding{51}\\ 
Pose driven &seen person             &\ding{51} &\ding{51} &\ding{51} &\ding{55} &\ding{55}&\ding{55}&\ding{55}&\ding{51} &\ding{51}\\ 
            &unseen person           &\ding{55} &\ding{55} &\ding{55} &\ding{55} &\ding{55}&\ding{55}&\ding{55}&\ding{51} &\ding{51}\\ 
\hline
\end{tabular}
\end{center}
\end{table*}






\section{Implementation Details}

\subsection{Baselines}
\textbf{NeuralBody~\cite{peng2021neural_human_nerf_2}}.
We use the officially released code and 
pre-trained model \footnote{https://github.com/zju3dv/neuralbody} for the experiments on the ZJU-MoCap dataset.

\textbf{Neural Human Performer~\cite{kwon2021neural_gene_human1}}.
We use the officially released code\footnote{https://github.com/YoungJoongUNC/Neural\_Human\_Performer} and pre-trained models for the experiments on the ZJU-MoCap dataset, and apply these models directly on DeepCap and DynaCap for testing the cross-dataset generalization.

\textbf{Animatable Nerf~\cite{peng2021animatable}}.
We use the officially released code\footnote{https://github.com/zju3dv/animatable\_nerf} and train the models on the ZJU-MoCap dataset.

\textbf{Neural Actor~\cite{liu2021neuralactor}}.
The quantitative and qualitative results shown in the paper are provided by the authors of Neural Actor.

\textbf{Keypoint NeRF~\cite{kpnerf}}.
We use the officially released code\footnote{https://github.com/facebookresearch/KeypointNeRF} and pre-trained models for the experiments on the ZJU-MoCap dataset, and apply these models directly on DeepCap and DynaCap for testing the cross-dataset generalization.

\textbf{MPS-NeRF~\cite{gao2022mps}}.
We use the officially released code\footnote{https://github.com/gaoxiangjun/MPS-NeRF} and the authors provide us codes to train it ZJU-MoCap dataset. 

\subsection{Network Structure}

\begin{figure*}[t!]
    \includegraphics[width=\textwidth]{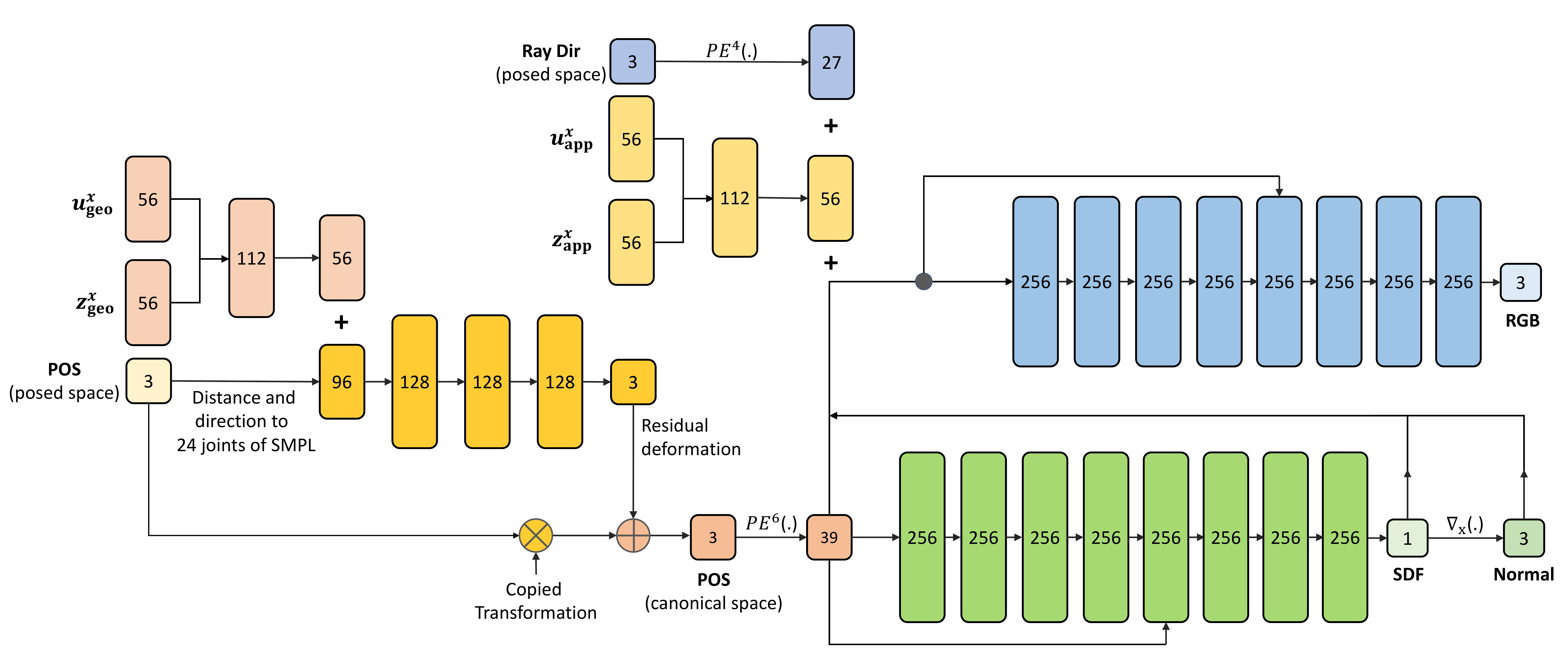}
    \caption{A visualization of the network architecture of the proposed deformation field, SDF field and color field. The number inside each block represents the vector’s dimension. "+" denotes  vector concatenation and the positional encoding function is defined as  PE$^L(x) = [x,\sin(2^0x),\cos(2^0x),...,\sin(2^{L-1}x),\cos(2^{L-1}x)]$. "$\nabla_x$" denotes a vector differential operator.}\label{fig:network_structure}
\end{figure*}

\textbf{Image feature extractor}
As briefly mentioned in the main paper, the image feature extractor CNN takes an input image $I^K \in \mathbb{R}^{H \times W \times 4} $ with its foreground mask as the last channel. The CNN has three down-sampling layers, and the output feature maps have the shapes of:
\begin{align}
    \{8 \times H \times W, \quad 16 \times H/2 \times W/2, \quad 32 \times H/4 \times W/4 \}
\end{align}
These feature maps are bilinearly upsampled to the highest resolution, and concatenated into a shape of $56 \times H \times W$.


\textbf{Graph convolutional networks}
We improve the standard graph convolution layer using an Encode-Process-Decode architecture proposed in ~\cite{pfaff2020mesh_gcn,sanchez2020gcn_simulate}. The encoder and decoder are Leaky-ReLU activated two-hidden-layer MLPs with LayerNorm. The processor consists of three identical message passing blocks, each containing two graph convolution layers~\cite{GCN_original} with BatchNorm. The edge features are scattered to the node features to fit in the training paradigm of GCN~\cite{GCN_original} and the graph topology keeps the same during message passing. All the layer sizes are set to 56.

\textbf{Neural fields}
Our method consists of 3 neural fields: (1) a deformation field, (2) a signed distance(SDF) field, and (3) a color field. All these fields are implemented as neural networks. The network architecture is shown in \ref{fig:network_structure}. The SDF field is modeled by a MLP consists of 8 hidden layers with the hidden size of 256. Following NeuS~\cite{wang2021_neus}, we use Softplus with $\beta = 100$ as the activation functions for all the hidden layers, and a skip connection is used to connect the input with the output of the fourth layer. The color field is modeled by another MLP sharing the same architecture with the SDF field, except that we use ReLu as the color field's activation functions for all the hidden layers. Positional encoding~\cite{mildenhall2020nerf_nerfbase} is applied to the spatial location $p$ with 6 frequencies and to the view direction $v$ with 4 frequencies. The input to the SDF field is the spatial location $p$, while the color field takes the spatial location $p$, the view direction $v$, the normal vector $n = \nabla f(p)$, and the appearance features $F_{app}^s,F_{app}^p$  as input. As the same as NeuS, we use weight normalization~\cite{salimans2016weight_norm} to stabilize the training process.

\textbf{Other hyperparameters}
For any spatial point, we query its 8 nearest vertices on SMPL to calculate its surface features $\bm{F}_*^s$. The bias term $B_*$ in the occlusion-aware self-attention mechanism is set to $-10$.

\section{Societal Impact}
In this section, we discuss the potential societal impact of this work. Our method enables the creation of an animatable human model from a sparse set of multi-view imagery of a person. It provides an easy way for users to create digital content using their own images and can be adopted to realize immersive AR and VR applications, such as telepresence. Our approach performs the best when the images are obtained from a cooperating subject in a controlled environment. Although our method is based on technical considerations, it still bears the risk that the human models of people are created and misused without consent. We urge our users to use this technique wisely only in the proper scenarios. We also hope that our method can contribute to detecting such fake content.



\bibliographystyle{abbrv-doi-hyperref}
\bibliography{egbib}